\documentclass[journal,final,twocolumn,lettersize,twoside]{IEEEtran}
\IEEEoverridecommandlockouts

\usepackage{amsfonts,amstext,amsmath,amssymb,amsthm}
\usepackage{color}
\usepackage{mathrsfs}
\usepackage{array}
\usepackage[caption=false,font=normalsize,labelfont=sf,textfont=sf]{subfig}
\usepackage{textcomp}
\usepackage{stfloats}
\usepackage{url}
\usepackage{verbatim}
\usepackage{graphicx}
\usepackage{cite}
\usepackage{euscript}       
\usepackage{bbm}

\theoremstyle{plain}
\newtheorem{theorem}{Theorem}

\newcommand{\ind}[1]{\mathbbm{1}_{#1}}   
\newcommand{\Exp}{{\mathsf{E}}}

\newcommand{\D}{\mathsf{D}}
\newcommand{\F}{\mathsf{F}}

\newcommand{\U}{\mathsf{U}}

\newcommand{\J}{\mathsf{J}}
\newcommand{\Q}{\mathsf{Q}}
\newcommand{\R}{\mathsf{R}}
\newcommand{\Z}{\mathsf{Z}}
\renewcommand{\H}{\mathsf{H}}
\renewcommand{\P}{\mathsf{P}}
\newcommand{\f}{\mathsf{f}}
\newcommand{\g}{\mathsf{g}}

\newcommand{\p}{\mathsf{p}}
\newcommand{\q}{\mathsf{q}}
\newcommand{\h}{\mathsf{h}}

\renewcommand{\a}{\mathsf{a}}
\renewcommand{\b}{\mathsf{b}}
\renewcommand{\c}{\mathsf{c}}
\renewcommand{\d}{\mathsf{d}}
\renewcommand{\L}{\mathsf{L}}
\renewcommand{\r}{\mathsf{r}}
\renewcommand{\u}{\mathsf{u}}
\newcommand{\uo}{\mathsf{u_o}}

\newcommand{\cI}{\mathcal{I}}

\newcommand{\cF}{\EuScript{F}}

\newcommand{\cQ}{\EuScript{Q}}

\newcommand{\cR}{\EuScript{R}}

\newcommand{\cW}{\EuScript{W}}
\newcommand{\cY}{\EuScript{Y}}
\newcommand{\cX}{\EuScript{X}}
\newcommand{\cU}{\EuScript{U}}
\newcommand{\cS}{\EuScript{S}}

\newcommand{\Real}{\mathbb{R}}

\begin{document}

\title{Data-Driven Estimation of Conditional Expectations, Application to Optimal Stopping and Reinforcement Learning} 

%
%

\author{\IEEEauthorblockN{George V. Moustakides, Emeritus Professor}\\
\IEEEauthorblockA{\textit{Department of Electrical and Computer Engineering} \\
University of Patras, Rion, GREECE\\
moustaki@upatras.gr}
}

\maketitle

\begin{abstract}
When the underlying conditional density is known, conditional expectations can be computed analytically or numerically. When, however, such knowledge is not available and instead we are given a collection of training data, the goal of this work is to propose simple and purely data-driven means for estimating directly the desired conditional expectation. Because conditional expectations appear in the description of a number of stochastic optimization problems with the corresponding optimal solution satisfying a system of nonlinear equations,
we extend our data-driven method to cover such cases as well. We test our methodology by applying it to Optimal Stopping and Optimal Action Policy in Reinforcement Learning.
\end{abstract}

\thispagestyle{empty}

\section{Introduction}
Conditional expectations appear in a multitude of stochastic optimization problems as part of defining the corresponding optimal solutions. Characteristic examples constitute Optimal Stopping, Reinforcement Learning, Optimal Control etc. In all these cases the identification of the desired solution requires exact knowledge of the underlying data probability density which in most practical applications is not available. 
Nowadays with the existence of great volumes of data one may wonder whether it is possible to develop data-driven methods to solve such problems based on ideas borrowed from Machine Learning. This is exactly the goal of our present work. In particular, we intend to offer computational means for identifying the desired solution by first introducing a method capable of estimating conditional expectations of known functions. Our technique will not be based on some form of initial probability density estimation but will directly identify the conditional expectation of interest.

Since our intention is to employ Machine Learning methods, we recall that these techniques mostly employ neural networks which are trained with available data. The ``design'' of networks is achieved through the solution of well-defined optimization problems involving \textit{expectations}. With the help of the Law of Large Numbers, expectations are being replaced by the existing data thus giving rise to data-driven techniques. For this reason in the sequel, our main effort focuses in proposing such optimization problems that are suitable for training and demonstrating that they indeed lead to the estimation of the functions of interest.

Before proceeding with the technical part we would like to point out that the methodology we are going to introduce is an extension of results presented in \cite{MB} and addressing the problem of likelihood ratio function estimation. This problem will also be mentioned here under the light of the richer experience we have accumulated.

\section{A General Optimization Problem}\label{sec:2}
Let us begin our presentation by introducing a simple optimization problem which will serve as the basis for the final, data-driven counterpart. Suppose that $\cX$ is a random vector and consider three scalar functions $\a(X),\b(X),\u(X)$ with $\a(X)>0$ that depend on the vector $X$ and two additional scalar functions $\phi(z),\psi(z)$ that depend on the scalar $z$. Fix $\a(X),\b(X),\phi(z),\psi(z)$ and for each $\u(X)$ define the following \textit{average cost}
\begin{equation}
\mathsf{J}(\mathsf{u})=\Exp_{\cX}\big[\a(\cX)\phi\big(\mathsf{u}(\cX)\big)+\b(\cX)\psi\big(\mathsf{u}(\cX)\big)\big],
\label{eq:cost0}
\end{equation}
where $\Exp_{\cX}[\cdot]$ denotes expectation with respect to $\cX$. We are interested in identifying the function $\uo(X)$ that solves the following optimization problem
\begin{equation}
\min_{\u(X)}\mathsf{J}(\u)=\min_{\u(X)}\Exp_{\cX}\big[\a(\cX)\phi\big(\mathsf{u}(\cX)\big)+\b(\cX)\psi\big(\mathsf{u}(\cX)\big)\big].
\label{eq:optim0}
\end{equation}
Solving \eqref{eq:optim0} can in general be challenging but if we limit ourselves to a particular class of functions $\phi(z),\psi(z)$ it is possible to come up with an explicit and straightforward answer. The theorem that follows specifies this class and the corresponding optimal solution.

\begin{theorem}\label{th:1}
For real $z$ let $\omega(z)$ be a strictly increasing scalar function and denote with $\cI$ its range of values. Select a second function $\rho(z)$ which is strictly negative and define two additional scalar functions $\phi(z),\psi(z)$ through their derivatives
\begin{equation}
\psi'(z)=\rho(z),~~~~~\phi'(z)=-\omega(z)\rho(z). 
\label{eq:th1-1}
\end{equation}
If $\mathrm{range}\big(\frac{\b(X)}{\a(X)}\big)\subseteq \cI$
then the optimal solution $\uo(X)$ of the optimization problem defined in \eqref{eq:optim0} is unique and satisfies
\begin{equation}
\omega\big(\uo(X)\big)=\frac{\b(X)}{\a(X)}. \label{eq:th1-2}
\end{equation}
\end{theorem}

\begin{IEEEproof}
Since the function $\u(X)$ which we would like to optimize depends on $X$ and also the cost is a result of averaging over $\cX$, it turns out that the minimization can be performed point-wise by interchanging expectation and optimization
\begin{multline*}
\min_{\u(X)}\Exp_{\cX}\big[\a(\cX)\phi\big(\mathsf{u}(\cX)\big)+\b(\cX)\psi\big(\mathsf{u}(\cX)\big)\big]\\
=\Exp_{\cX}\left[\a(\cX)\min_{\u(\cX)}\Big\{\phi\big(\mathsf{u}(\cX)\big)+\r(\cX)\psi\big(\mathsf{u}(\cX)\big)\Big\}\right],
\end{multline*}
where for simplicity we denote $\r(X)=\frac{\b(X)}{\a(X)}$. The above equality is true because by assumption $\a(X)>0$. We must emphasize that in general changing the order of expectation and minimization results in an inequality. However here because the minimization is with respect to the function $\u(X)$ that depends only on the quantity that is averaged, it is straightforward to show that we actually enjoy exact equality. The minimization
$$
\min_{\u(X)}\left\{\phi\big(\u(X)\big)+\r(X)\psi\big(\u(X)\big)\right\}
$$
can now be performed for \textit{each individual} $X$ (point-wise).
Fixing $X$ means that $\u(X)$ becomes a scalar quantity $\u$ while the ratio $\r(X)$ becomes a scalar number $\r$. Consequently for each $\r\in\cI$ we need to perform a minimization with respect to $\u$ of the form
$$
\min_{\u}\left\{\phi(\u)+\r\psi(\u)\right\}.
$$
Taking the derivative of $\phi(\u)+\r\psi(\u)$ with respect to $\u$ and using the definition of the two functions from \eqref{eq:th1-1} yields
$$
\phi'(\u)+\r\psi'(\u)=\big(\r-\omega(\u)\big)\rho(\u).
$$
Because of the strict increase of $\omega(\cdot)$ and the strict negativity of $\rho(\cdot)$ we conclude that the derivative is negative for $\u$ satisfying $\omega(\u)<\r$ and positive for $\omega(\u)>\r$ which implies that for $\u=\uo$ satisfying $\omega(\uo)=\r$ we have a unique global minimum. Since this is true for every $X$ we deduce that the optimal function $\uo(X)$ is such that $\omega\big(\uo(X)\big)=\r(X)=\frac{\b(X)}{\a(X)}$. This concludes the proof.
\end{IEEEproof}

Because $\omega(z)$ is strictly increasing the set $\cI$ can be either the whole real line $\Real$ or a semi-infinite interval $(a,\infty)$, $(-\infty,a)$ or a finite interval $(a,b)$. Instead of open we can have closed intervals as well.

\subsection{Versions of the Optimization Problem}\label{sec:2A}
Selecting various pairs of functions $\a(X),\b(X)$ in combination with the probability density of the random vector $\cX$ produces interesting and practically meaningful optimization problems. First we address the problem of likelihood ratio identification of two densities, which is considered in detail in \cite{MB}. 

\subsubsection{Identification of Likelihood Ratios} Let $\g(X),\f(X)$ be two possible densities for the random vector $\cX$ and assume that $\f(X)=0$ when $\g(X)=0$ (to avoid unbounded ratios). For any scalar function $\u(X)$ define the cost
\begin{equation}
\J(\u)=\Exp_{\cX}^{\g}\big[\phi\big(\u(\cX)\big)\big]+\Exp_{\cX}^{\f}\big[\psi\big(\u(\cX)\big)\big]
\label{eq:optimLR1}
\end{equation}
where $\Exp_{\cX}^{\g}[\cdot],\Exp_{\cX}^{\f}[\cdot]$ denote expectation with respect to $\cX$ under the densities $\g(X),\f(X)$ respectively. Applying a simple change of measure and denoting with $\L(X)=\frac{\f(X)}{\g(X)}$ the likelihood ratio of the two densities we have
\begin{equation}
\J(\u)=\Exp_{\cX}^{\g}\big[\phi\big(\u(\cX)\big)+\L(\cX)\psi\big(\u(\cX)\big)\big]
\label{eq:optimLR2}
\end{equation}
which is under the form of \eqref{eq:cost0}. This suggests that for the minimization of \eqref{eq:optimLR1} we can write
\begin{multline*}
\min_{\u(X)}\J(\u)=\min_{\u(X)}\Big\{\Exp_{\cX}^{\g}\big[\phi\big(\u(\cX)\big)\big]+\Exp_{\cX}^{\f}\big[\psi\big(\u(\cX)\big)\big]\Big\}\\
=\min_{\u(X)}\Exp_{\cX}^{\g}\big[\phi\big(\u(\cX)\big)+\L(\cX)\psi\big(\u(\cX)\big)\big].
\end{multline*}
Application of Theorem\,\ref{th:1} produces as optimal solution the function $\uo(X)$ that satisfies $\omega\big(\uo(X)\big)=\L(X)$. Consequently we identify the likelihood ratio function of the two densities through an optimization involving expectations.

\subsubsection{Identification of Ratio of Conditional Densities}
We can extend the previous result to cover the likelihood ratio of \textit{conditional densities}. Specifically let the pair of random vectors $(\cY,\cX)$ be described by two possible joint densities $\g(Y,X),\f(Y,X)$. Write $\g(Y,X)=\g(Y|X)\g(X), \f(Y,X)=\f(Y|X)\f(X)$ and as above denote $\L(X)=\frac{\f(X)}{\g(X)}$ the likelihood ratio of the two marginals. For a scalar function $\u(Y,X)$ consider the following cost
\begin{equation}
\J(\u)=\Exp_{\cY,\cX}^{\g}\big[\L(\cX)\phi\big(\u(\cY,\cX)\big)\big]+\Exp_{\cY,\cX}^{\f}\big[\psi\big(\u(\cY,\cX)\big)\big].
\label{eq:optim_cLR}
\end{equation}
Applying again a change of measure we can write
$$
\J(\u)=\Exp_{\cY,\cX}^{\g}\left[\L(\cX)\phi\big(\u(\cY,\cX)\big)+\frac{\f(\cY,\cX)}{\g(\cY,\cX)}\psi\big(\u(\cY,\cX)\big)\right],
$$
which, according to Theorem\,\ref{th:1} when minimized over $\u(Y,X)$ will yield an optimal solution of the form
$$
\omega\big(\uo(Y,X)\big)=\frac{\f(Y,X)}{\g(Y,X)}\frac{1}{\L(X)}=\frac{\f(Y|X)}{\g(Y|X)}.
$$
Of course $\L(X)$ as we have seen in the previous case can be obtained by optimizing a cost of the form of \eqref{eq:optimLR1} which in turn can be used in the optimization of \eqref{eq:optim_cLR} to identify the likelihood ratio of the conditional densities.

\subsubsection{Identification of Conditional Expectations}
Let us now address the main problem of interest. We must point out that the advantage of the method we are going to introduce is that we can estimate \textit{conditional expectations} by solving optimization problems involving \textit{regular expectations}. This is particularly useful from a practical point of view since as we will see in the next section, it is straightforward to solve such problems under a data-driven setup.

Consider two scalar functions $\c(Y)$ and $\d(Y)$ with $\c(Y)>0$. For the pair of random vectors $(\cY,\cX)$ define the conditional expectations $\a(X)=\Exp_{\cY}[\c(\cY)|\cX=X], \b(X)=\Exp_{\cY}[\d(\cY)|\cX=X]$. We clearly have $\a(X)>0$. For a scalar function $\u(X)$ we now define the cost
\begin{equation}
\J(\u)=\Exp_{\cY,\cX}\big[\c(\cY)\phi\big(\u(\cX)\big)+\d(\cY)\psi\big(\u(\cX)\big)\big],
\label{eq:optim_ce}
\end{equation}
where expectation is with respect to \textit{the pair} $(\cY,\cX)$. Note that even though the expectation is with respect to $(\cY,\cX)$ the function we would like to optimize depends \textit{only} on $X$.
Using the tower property of expectation the cost can be rewritten as
\begin{align*}
\J(\u)&=\Exp_{\cX}\Big[\Exp_{\cY}\big[\c(\cY)|\cX\big]\phi\big(\u(\cX)\big)+\Exp_{\cY}\big[\d(\cY)|\cX\big]\psi\big(\u(\cX)\big)\Big]\\
&=\Exp_{\cX}\Big[\a(\cX)\phi\big(\u(\cX)\big)+\b(\cX)\psi\big(\u(\cX)\big)\Big].
\end{align*}
Consequently, minimizing $\J(\u)$ defined in \eqref{eq:optim_ce} over $\u(X)$, by application of Theorem\,\ref{th:1}, yields
$$
\omega\big(\uo(X)\big)=\frac{\b(X)}{\a(X)}=\frac{\Exp_{\cY}[\d(\cY)|\cX=X]}{\Exp_{\cY}[\c(\cY)|\cX=X]}.
$$
In other words it computes the ratio of the two conditional expectations. If we select $\c(Y)=1$ then
$$
\omega\big(\uo(X)\big)=\Exp_{\cY}[\d(\cY)|\cX=X],
$$
namely we directly identify the conditional expectation of $\d(\cY)$ with respect to $\cY$ given $\cX=X$ as the solution of an optimization problem involving \textit{regular} instead of conditional expectations.

This result can be further extended by considering two different densities $\g(Y,X),\f(Y,X)$ and for a function $\u(X)$ define the cost
\begin{equation}
\J(\u)=\Exp^{\g}_{\cY,\cX}\big[\c(\cY)\phi\big(\u(\cX)\big)\big]+\Exp^{\f}_{\cY}\big[\d(\cY)\psi\big(\u(\cX)\big)\big].
\label{eq:optim_ce2}
\end{equation}
In order to specify the optimizer of $\J(\u)$ we observe that
\begin{multline*}
\!\!\!\!\J(\u)=\Exp^{\g}_{\cX}\Big[\Exp^{\g}_{\cY}\big[\c(\cY)|\cX\big]\phi\big(\u(\cX)\big)\Big]+\Exp^{\f}_{\cX}\Big[\Exp^{\f}_{\cY}\big[\d(\cY)|\cX\big]\psi\big(\u(\cX)\big)\Big]\\
=\Exp^{\g}_{\cX}\Big[\Exp^{\g}_{\cY}\big[\c(\cY)|\cX\big]\phi\big(\u(\cX)\big)+\L(\cX)\Exp^{\f}_{\cY}\big[\d(\cY)|\cX\big]\psi\big(\u(\cX)\big)\Big].
\end{multline*}
where $\L(X)=\frac{\f(X)}{\g(X)}$ is the likelihood ratio of the two marginal densities.
Applying Theorem\,\ref{th:1} we conclude that $\J(\u)$ is optimized by
$$
\omega\big(\uo(X)\big)=\L(X)\frac{\Exp^{\f}_{\cY}\big[\d(\cY)|\cX=X\big]}{\Exp^{\g}_{\cY}\big[\c(\cY)|\cX=X\big]}.
$$
Note that the advantage of this approach is that we end up with a \textit{single} function $\uo(X)$ that identifies the ratio of the two conditional expectations. An alternative idea would be to identify the two conditional expectations of the previous ratio separately through two different optimization problems which is clearly not as efficient.

\subsection{Examples of functions $\omega(z),\rho(z),\phi(z),\psi(z)$}\label{ssec:2.A}
According to Theorem\,\ref{th:1}, a very important quantity in selecting the two functions $\omega(z),\rho(z)$ is the range of values $\cI$ of $\omega(z)$ which must cover the range of values of the ratio $\frac{\b(X)}{\a(X)}$. The latter may not be exactly known but, as stated in the theorem, it is sufficient that $\cI$ is a superset of this range. We will provide examples for three different cases of $\cI$ namely $\cI=\Real$, $\cI=[a,\infty)$ and $\cI=(a,b)$ with $a<b$. The first covers the case where the range is completely unknown while the second and third refer to cases with partially known range. For example if the function $\d(Y)$ for which we would like to compute the conditional expectation is nonnegative or between $a$ and $b$ then the same property holds true for its conditional expectation.
\newpage
\noindent \underline{[A].~~$\cI=\Real$}
\vskip0.2cm
\noindent[A1].~~$\omega(z)=z$ and $\rho(z)=-1$, results in 
$$
\phi(z)=\frac{z^2}{2}, ~~~~\psi(z)=-z.
$$
This particular selection is the most popular in practice and it is known as the Mean Square Error (MSE) criterion.
\vskip0.2cm
\noindent[A2].~~$\omega(z)=\sinh(z)$ and $\rho(z)=-e^{-0.5|z|}$ results in
\begin{align*}
\phi(z)&=(e^{0.5|z|}-1)+\frac{1}{3}(e^{-1.5|z|}-1),\\
\psi(z)&=2\text{sign}(z)(e^{-0.5|z|}-1).
\end{align*}
\vskip0.2cm
\noindent[A3].~~$\omega(z)=\text{sign}(z)\big(e^{|z|}-1\big)$ and $\rho(z)=-e^{-0.5|z|}$, results in
$$
\phi(z)=4\cosh(0.5z), ~~~~\psi(z)=2\text{sign}(z)(e^{-0.5|z|}-1).
$$
\vskip0.2cm
\noindent \underline{[B].~~$\cI=(a,\infty)$}
\vskip0.2cm
\noindent[B1].~~$\omega(z)=a+e^z$ and $\rho(z)=-\frac{1}{1+e^z}$, results in
$$
\phi(z)=a\log(1+e^{-z})+\log(1+e^z), ~~~~\psi(z)=\log(1+e^{-z}).
$$
This version when $a=0$ resembles the \textit{cross entropy} method introduced in \cite{GAN} for the design of GANs where they employ $\phi(z)=-\log(\frac{1}{z})$, $\psi(z)=-\log(\frac{1}{1-z})$ with $z\in(0,1)$. We can see that we propose the same functions but with $z$ replaced in our approach by $\frac{1}{1+e^z}$ and $z\in\Real$.
\vskip0.1cm
\noindent[B2].~~$\omega(z)=a+e^z$ and $\rho(z)=-e^{-0.5 z}$, results in
$$
\phi(z)=-2ae^{-0.5z}+2e^{0.5z}, ~~~~\psi(z)=2e^{-0.5z}.
$$

\noindent\textbf{Remark~1.}~We note that when [B1] or [B2] is applied in the problem of likelihood ratio identification with $a=0$ then
$$
\omega\big(\uo(X)\big)=e^{\uo(X)}=\frac{\f(X)}{\g(X)}~\Rightarrow
\uo(X)=\log\left(\frac{\f(X)}{\g(X)}\right),
$$
suggesting that the optimal function $\uo(X)$ is equal to the \textit{log-likelihood ratio} of the two densities which, in problems as hypothesis testing, is often more convenient to use than the likelihood ratio itself. 

In case the range is $\cI=(-\infty,a)$ we simply consider the conditional expectation of the function $-\d(Y)$ which will have a range of the form $\cI=(-a,\infty)$.
\vskip0.2cm
\noindent \underline{[C].~~$\cI=(a,b)$}
\vskip0.2cm
\noindent[C1].~~$\omega(z)=a\frac{1}{1+e^z}+b\frac{e^z}{1+e^z}$ and $\rho(z)=-\frac{e^z}{1+e^z}$, results in
$$
\phi(z)=\frac{b-a}{1+e^z}+b\log(1+e^z), ~~~~\psi(z)=-\log(1+e^{z}).
$$
\vskip0.2cm
\noindent[C2].~~$\omega(z)=a\frac{1}{1+e^z}+b\frac{e^z}{1+e^z}$ and $\rho(z)=-e^{-z}$, results in
$$
\phi(z)=(b-a)\log\Big(\frac{e^z}{1+e^z}\Big)-ae^{-z}, ~~~~\psi(z)=e^{-z}.
$$
Function $\omega(z)$ reduces to the classical sigmoid for the (0,1) interval.
\vskip0.2cm
\noindent\textbf{Remark~2.}~It is straightforward to propose alternative combinations of $\omega(z),\rho(z)$ that satisfy the requirements of Theorem\,\ref{th:1}. We also need to point out that for any range $\cI$ of interest it is always possible to employ a pair which is designed for a wider range. For example we can apply [A1] (MSE) in the case of functions with range in an interval $(a,b)$ instead of the suggested [C1], [C2]. In fact this is common practice in the literature.

\section{Data-Driven Estimation}
Solving the optimization problem defined in \eqref{eq:optim0} in order to identify the optimal function $\uo(X)$ requires knowledge of the underlying probability density of $\cX$. Our goal in the analysis that follows is to relax this requirement and replace it with the existence of a number of realizations of $\cX$. This constitutes the data-driven version of the problem.

The first classical step we adopt in the direction of a data-driven approach is to replace the unknown function $\u(X)$ with a parametric family $\u(X,\theta)$ involving a \textit{finite} set of parameters $\theta$. Of course we require this family to enjoy the \textit{universal approximation property}, namely to have the ability to approximate arbitrarily close any sufficiently smooth function provided we select a large enough model. This property is guaranteed for neural networks (shallow or deep) according to \cite{CYB,HTW} but it may be enjoyed by other parametric classes as well. In the sequel we limit ourselves to neural networks but similar conclusions can be claimed for any other such class. 

Let us now replace $\u(X)$ with a neural network $\u(X,\theta)$, then the cost function in \eqref{eq:cost0} becomes
\begin{equation}
\mathsf{J}(\theta)=\Exp_{\cX}\big[\a(\cX)\phi\big(\mathsf{u}(\cX,\theta)\big)+\b(\cX)\psi\big(\mathsf{u}(\cX,\theta)\big)\big],
\label{eq:cost10}
\end{equation}
depending only on the parameters of the network, while the corresponding optimization in \eqref{eq:optim0} is replaced by
\begin{equation}
\min_{\theta}\mathsf{J}(\theta)=\min_{\theta}\Exp_{\cX}\big[\a(\cX)\phi\big(\mathsf{u}(\cX,\theta)\big)+\b(\cX)\psi\big(\mathsf{u}(\cX,\theta)\big)\big].
\label{eq:optim10}
\end{equation}
For sufficiently large neural network if $\theta_{\mathsf{o}}$ is the optimizer of \eqref{eq:optim10} we expect the corresponding neural network $\u(X,\theta_{\mathsf{o}})$ to satisfy $\u(X,\theta_{\mathsf{o}})\approx\uo(X)$ with the latter being the optimizer of the original problem in \eqref{eq:optim0}. In other words, with the optimal finite dimensional version of the problem defined in \eqref{eq:optim10} we approximate the optimal function solving its infinite dimensional counterpart in~\eqref{eq:optim0}.
As in the original problem in \eqref{eq:optim0}, the finite dimensional version \eqref{eq:optim10} is defined in terms of the pdf of $\cX$. Therefore, let us now assume that we are under a data-driven setup. 

As we can see the cost in \eqref{eq:cost10} can be put under the general form
$$
\J(\theta)=\Exp_{\cX}[\h(\cX,\theta)],
$$
where $\h(X,\theta)$ is a scalar deterministic function and \eqref{eq:optim10} corresponds to solving the optimization problem
\begin{equation}
\min_{\theta}\J(\theta)=\min_{\theta}\Exp_{\cX}[\h(\cX,\theta)],
\label{eq:gopt}
\end{equation}
for the case where we have a set of realization $\{X_1,\ldots,X_n\}$ (training set)  of $\cX$ which replaces the exact knowledge of the probability density function of $\cX$.

The obvious possibility is to evoke the classical Law of Large Numbers (LLN) and approximate the expectation by the following data-driven cost 
$$
\hat{\J}(\theta)=\frac{1}{n}\sum_{i=1}^n\h(X_i,\theta)\approx\J(\theta).
$$
Cost $\hat{\J}(\theta)$ is a completely known function of $\theta$ and can therefore be minimized with the help, for example, of the Gradient Descent (GD) iterative algorithm
\begin{equation}
\theta_t=\theta_{t-1}-\mu\sum_{i=1}^n\nabla_{\!\!\theta}\h(X_i,\theta_{t-1})
\label{eq:LLNs}
\end{equation}
where $\mu>0$ is the step-size (learning rate) and $\nabla_{\!\!\theta}\h(X,\theta)$ is the gradient with respect to $\theta$ of the scalar function $\h(X,\theta)$. Note that we have absorbed in $\mu$ the division by the constant $n$. As we can see, the GD in each iteration requires the computation of the gradients of all realizations which could be computationally demanding especially when the training set is large.

An alternative approach would be to employ the Stochastic Gradient Descent (SGD) algorithm which consists in applying the update
\begin{equation}
\theta_t=\theta_{t-1}-\mu\nabla_{\!\!\theta}\h(X_t,\theta_{t-1})
\label{eq:SGDD}
\end{equation}
where we retain only a \textit{single} gradient evaluated for a realization $X_t$ from the training set. In each iteration we use a different data point and when all realization are exhausted (epoch) then we reuse them starting from the beginning of the training set (after possibly randomly shuffling the data). 

There is also the version where in each iteration we employ micro-blocks of $m$ elements from the training set and replace the single gradient of the classical version with the average of the gradients computed over the block (again division by $m$ is absorbed in $\mu$)
$$
\theta_t=\theta_{t-1}-\mu\sum_{i=1}^m\nabla_{\!\!\theta}\h(X_{(t-1)m+i},\theta_{t-1}).
$$
The SGD is clearly computationally less demanding per iteration than the GD. 

Relating \eqref{eq:SGDD} to the solution of the original problem defined in \eqref{eq:gopt} is not as straightforward as in \eqref{eq:LLNs} where we simply call upon the LLN. A more sophisticated Stochastic Approximation theory \cite{BMP} is required to demonstrate that the SGD version indeed provides the desired estimate. Furthermore, as it has been observed in practice, the behavior of the two algorithms (GD vs SGD)  with respect to convergence speed and capability to avoid undesirable local minima can be quite different.

\subsection{Data-Driven Likelihood Ratio Estimation}
In the previous section when we discussed the problem of likelihood ratio identification we considered the existence of two densities $\g(X),\f(X)$ and the need to minimize the performance measure in \eqref{eq:optimLR1} with respect to the unknown function $\u(X)$. 

Following similar steps as the ones described above, we first replace $\u(X)$ with a neural network $\u(X,\theta)$ then, instead of the two densities we assume existence of two datasets $\{X_1^{\g},\ldots,X_{n_{\g}}^{\g}\}$ and  $\{X_1^{\f},\ldots,X_{n_{\f}}^{\f}\}$ sampled from $\g(X),\f(X)$ respectively. Using the definitions in \eqref{eq:th1-1} and denoting with $\nabla_{\!\!\theta}\u(X,\theta)$ the gradient with respect to $\theta$ of the neural network $\u(X,\theta)$ we can write for the GD iteration
\begin{multline*}
\hskip1cm\theta_t=\theta_{t-1}+\allowdisplaybreaks\\ \allowdisplaybreaks
\mu\Big\{
\frac{1}{n_{\g}}\sum_{i=1}^{n_{\g}}\omega\big(\u(X_i^{\g},\theta_{t-1})\big)\rho\big(\u(X_i^{\g},\theta_{t-1})\big)\nabla_{\!\!\theta}\u(X_i^{\g},\theta_{t-1})\allowdisplaybreaks\\ \allowdisplaybreaks
-\frac{1}{n_{\f}}\sum_{j=1}^{n_{\f}}\rho\big(\u(X_j^{\f},\theta_{t-1})\big)\nabla_{\!\!\theta}\u(X_j^{\f},\theta_{t-1})
\Big\}.\hskip1cm
\end{multline*}

For the SGD version we assume that the two datasets are randomly mixed with the samples retaining their labels (i.e. whether they come from $\g$ or $\f$). The samples are used one after the other and the update per iteration depends on the corresponding label of the sample. Specifically when $X_t$ is from $\mathsf{g}$ then
$$
\theta_t=\theta_{t-1}+
\frac{\mu}{n_{\g}}\omega\big(\u(X_t,\theta_{t-1})\big)\rho\big(\u(X_t,\theta_{t-1})\big)\nabla_{\!\!\theta}\u(X_t,\theta_{t-1}),
$$
whereas when $X_t$ is from $\mathsf{f}$ then
$$
\theta_t=\theta_{t-1}-
\frac{\mu}{n_{\f}}\rho\big(\u(X_t,\theta_{t-1})\big)\nabla_{\!\!\theta}\u(X_t,\theta_{t-1}).
$$
When $n_{\g}=n_{\f}$ we can simplify the SGD algorithm by not mixing the two datasets and by employing in each iteration one sample from each dataset as follows (division by $n_{\f}=n_{\g}$ is absorbed in $\mu$)
\begin{multline*}
\!\!\!\!\theta_t=\theta_{t-1}+\mu\Big\{\omega\big(\u(X_t^{\g},\theta_{t-1})\big)\rho\big(\u(X_t^{\g},\theta_{t-1})\big)\nabla_{\!\!\theta}\u(X_t^{\g},\theta_{t-1})\\
-\rho\big(\u(X_t^{\f},\theta_{t-1})\big)\nabla_{\!\!\theta}\u(X_t^{\f},\theta_{t-1})\Big\}.
\end{multline*}
Gradient type algorithms for estimating the likelihood ratio of conditional probability densities can be designed in a similar way.

\subsection{Data-Driven Estimation of Conditional Expectations}\label{ssec:ddece}
We would like to emphasize once more that the advantage of the proposed methodology is the fact that we can estimate conditional expectations by solving optimization problems involving regular expectations. Let us recall the cost in \eqref{eq:optim_ce} and consider the simplified version with $\c(Y)=1$. For a function $\d(Y)$ we are interested in estimating the conditional expectation $\Exp_{\cY}[\d(\cY)|\cX=X]$. Following our usual practice, the function $\u(X)$ is replaced by a neural network $\u(X,\theta)$ and the joint density $\f(Y,X)$ by a collection of pairs $\{(Y_1,X_1),\ldots,(Y_n,X_n)\}$ sampled from it.
The finite dimensional version of the cost function in \eqref{eq:optim_ce} then becomes
$$
\J(\theta)=\Exp_{\cY,\cX}\Big[\phi\big(\u(\cX,\theta)\big)+\d(\cY)\psi\big(\u(\cX,\theta)\big)\Big],
$$
which suggests the following data-driven cost
$$
\hat{\J}(\theta)=\frac{1}{n}\sum_{i=1}^n\Big\{\phi\big(\u(X_i,\theta)\big)+\d(Y_i)\psi\big(\u(X_i,\theta)\big)\Big\}
$$
and the corresponding GD algorithm for its minimization
\begin{multline}
\theta_t=\theta_{t-1}-
\mu\sum_{i=1}^n \Big\{\d(Y_i)-\omega\big(\u(X_i,\theta_{t-1})\big)\Big\}\times\\
\rho\big(\u(X_i,\theta_{t-1})\big)\nabla_{\!\!\theta}\u(X_i,\theta_{t-1}) .
\label{eq:GD}
\end{multline}
For the SGD we can write
\begin{multline}
\theta_t=\theta_{t-1}-\mu\Big\{\d(Y_t)-\omega\big(\u(X_t,\theta_{t-1})\big)\Big\}\times\\[2pt]
\rho\big(\u(X_t,\theta_{t-1})\big)\nabla_{\!\!\theta}\u(X_t,\theta_{t-1}).
\label{eq:SGD}
\end{multline}
When $\theta_t$ converges to $\theta_{\mathsf{o}}$, we expect that $\omega\big(\u(X,\theta_{\mathsf{o}})\big)\approx\Exp_{\cY}[\d(\cY)|\cX=X]$ without any knowledge of the joint or the conditional probability density of $\cY$ given $\cX$.
\vskip0.2cm
\noindent\textbf{Remark~3.}~As we can see from \eqref{eq:GD}, \eqref{eq:SGD} for the updates we require the two initial functions $\omega(z),\rho(z)$ and not $\phi(z),\psi(z)$. The latter are needed only for the computation of the corresponding cost $\hat{\J}(\theta_t)$ which can be used to monitor the stability and convergence of the iterations.
In fact, in order to experience stable updates in \eqref{eq:GD}, \eqref{eq:SGD} we need to ensure that $\d(Y)\in\cI$. In other words, the range of $\omega(z)$ must cover the range of values of $\d(Y)$. 

We should mention that in \eqref{eq:GD}, \eqref{eq:SGD} it is very common to employ the ADAM version \cite{ADAM} where gradient elements are normalized by the square root of their running power. This idea establishes a more uniform convergence for the components of the parameter vector. Powers are estimated with the help of exponential windowing with a forgetting factor $\lambda$.

The function $\omega(z)$ which is applied after the computation of the output of the neural network $\u(X,\theta_{\mathrm o})$ can be seen as an \textit{output activation} function. We should however emphasize that the updates in \eqref{eq:GD} or \eqref{eq:SGD} would have been different if we had considered $\omega(z)$ as part of the neural network from the start. Indeed the gradient with respect to the parameters, unlike in \eqref{eq:GD} and \eqref{eq:SGD}, would have also included the derivative $\omega'(z)$ which is now absent.

\subsection{Numerical Computation of Conditional Expectations}\label{ssec:numerical}
In order to evaluate the proposed data-driven estimation method we would need to compare its results with the exact conditional expectation, for characteristic examples. Since it is not always possible to analytically compute the conditional expectation we would like to offer a computational technique based on simple numerical integration rules.

For the pair $(\cY,\cX)$ where $\cY$ and $\cX$ are \textit{scalars} let $\f(Y|X)$ be the conditional pdf and $\F(Y|X)$ the corresponding conditional cdf. The conditional expectation $\Exp_{\cY}[\d(\cY)|\cX=X]$ for a known function $\d(Y)$ can then be written as
\begin{multline}
\u(X)=\Exp_{\cY}[\d(\cY)|\cX=X]\\[2pt]
=\int\d(Y)\f(Y|X)\,dY=\int\d(Y)\,d\F(Y|X).
\label{eq:num_int}
\end{multline}
If we sample $Y$ and $X$ over sufficiently large intervals at $\{Y_1,\ldots,Y_n\}$ and $\{X_1,\ldots,X_m\}$ respectively then we can generate the doubly indexed sequence $\F(Y_j|X_i)$ and the two sequences $\d(Y_j),\u(X_i)$ where $j=1,\ldots,n$ and $i=1,\ldots,m$. We note that the values $\u(X_i)$ are the samples of the conditional expectation we would like to determine. By averaging the forward and backward version of the rectangle rule in the second integral in \eqref{eq:num_int}, the conditional expectation can enjoy the following approximation
\begin{align*}
\u(X_i)&\approx\frac{1}{2}\sum_{j=1}^{n-1}\d(Y_j)[\F(Y_{j+1}|X_i)-\F(Y_j|X_i)]\allowdisplaybreaks\\ \allowdisplaybreaks
&\hskip1.5cm+\frac{1}{2}\sum_{j=2}^{n}\d(Y_j)[\F(Y_{j}|X_i)-\F(Y_{j-1}|X_i)]\allowdisplaybreaks\\ \allowdisplaybreaks
&=\d(Y_1)\frac{\F(Y_2|X_i)-\F(Y_1|X_i)}{2} \allowdisplaybreaks\\  \allowdisplaybreaks
&~~+\d(Y_2)\frac{\F(Y_3|X_i)-\F(Y_1|X_i)}{2}+\cdots \allowdisplaybreaks\\  \allowdisplaybreaks
&~~+\d(Y_{n-1})\frac{\F(Y_n|X_i)-\F(Y_{n-2}|X_i)}{2} \allowdisplaybreaks\\  \allowdisplaybreaks
&~~+\d(Y_n)\frac{\F(Y_n|X_i)-\F(Y_{n-1}|X_i)}{2}.
\end{align*}
We note that the first and last term in the last sum are different from the intermediate terms. The above formula can be conveniently rewritten as a matrix/vector product. Indeed if we define the vectors $\U=[\u(X_1),\ldots,\u(X_m)]^\intercal$, $\D=[\d(Y_1),\ldots,\d(Y_n)]^\intercal$ and the matrix $\cF$ of dimensions $m\times n$ with the $i$-th row of the matrix having the following elements $(\cF)_{i1}=0.5[\F(Y_2|X_i)-\F(Y_1|X_i)]$, $(\cF)_{ij}=0.5[\F(Y_{j+1}|X_i)-\F(Y_{j-1}|X_i)],j=2,\ldots,n-1$ and $(\cF)_{in}=0.5[\F(Y_n|X_i)-\F(Y_{n-1}|X_i)]$, then we can write $\U\approx\cF\,\D$, suggesting that the product $\cF\,\D$ provides an approximation to the sampled values of the conditional expectation.

In case we have available the conditional pdf $\f(Y|X)$ but it is not possible to compute analytically the conditional cdf $\F(Y|X)$ we can use the classical version of the forward and backward rectangle rule applied to the first integral in \eqref{eq:num_int}, namely
\begin{align*}
&\u(X_i)\approx\frac{1}{2}\sum_{j=1}^{n-1}\d(Y_j)\f(Y_j|X_i)(Y_{j+1}-Y_j)\\
&\hskip3cm+\frac{1}{2}\sum_{j=2}^{n}\d(Y_j)\f(Y_{j}|X_i)(Y_j-Y_{j-1})\\
&=\sum_{j=1}^{n-1}\frac{1}{2}\Big(\d(Y_j)\f(Y_j|X_i)
+\d(Y_{j+1})\f(Y_{j+1}|X_i)\Big)(Y_{j+1}-Y_j),
\end{align*}
leading to the trapezoidal rule. This can also be combined to the more convenient matrix/vector product by properly redefining the matrix $\cF$.

As mentioned, the numerical technique will be used whenever it is impossible to obtain an analytic formula and will serve as a point of reference for the proposed data-driven method. We should of course keep in mind that the numerical method requires exact knowledge of the conditional pdf (cdf) whereas the data-driven technique we developed relies only on training data. 

Even though the numerical method is presented for the scalar case it is possible to extend it to accommodate random vectors $\cY,\cX$. However, very quickly we realize that as we consider larger dimensions the amount of necessary computations increases dramatically with the method suffering from the ``curse of dimensionality''. 

When we apply this numerical method we have to make sure that the intervals we sample are such that the conditional probability mass which is left outside the sampled $Y$-interval is negligible for all sampled values of $X$ and so is the probability mass left outside the sampled $X$-interval. If this is not the case then changing the size of the intervals produces inconsistent and therefore questionable results.

\subsection{Examples}
Let us apply our idea to two examples. Consider $\cY,\cX$ scalar random variables related through the equations: a)~$\cY=\text{sign}(\cX)\,\cX^2+\cW$, and b)~$\cY=\ind{[-1,1]}(\cX+\cW)$, where $\ind{A}(X)$ denotes the indicator function of the set $A$. In both cases $\cX$ is standard Gaussian while $\cW$ plays to role of noise which we assume to be zero-mean Gaussian with variance $\sigma^2_{\cW}=0.1$. We are interested in estimating $\Exp_{\cY}[\cY|\cX=X]$ and it is straightforward to see that for a)~we have $\Exp_{\cY}[\cY|\cX=X]=\text{sign}(X)X^2$, whereas for b)~$\Exp_{\cY}[\cY|\cX=X]=\Phi(\frac{1-X}{\sigma_{\cW}})-\Phi(\frac{-1-X}{\sigma_{\cW}})$, where $\Phi(x)$ is the cdf of the standard Gaussian. Here we have an analytic formula for both conditional expectations but even if we had used the numerical method proposed in Section\,\ref{ssec:numerical} the two results would have been indistinguishable.

\begin{figure}[b!]
\centerline{\includegraphics[scale=0.48]{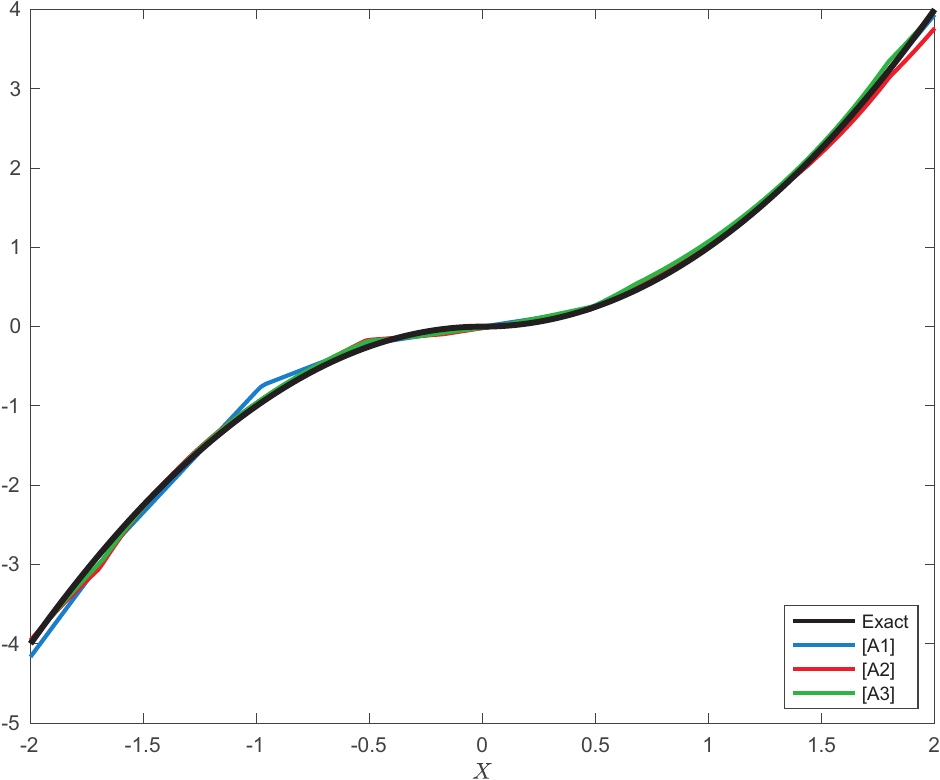}}
\centerline{(a)}
\vskip0.4cm
\centerline{\includegraphics[scale=0.48]{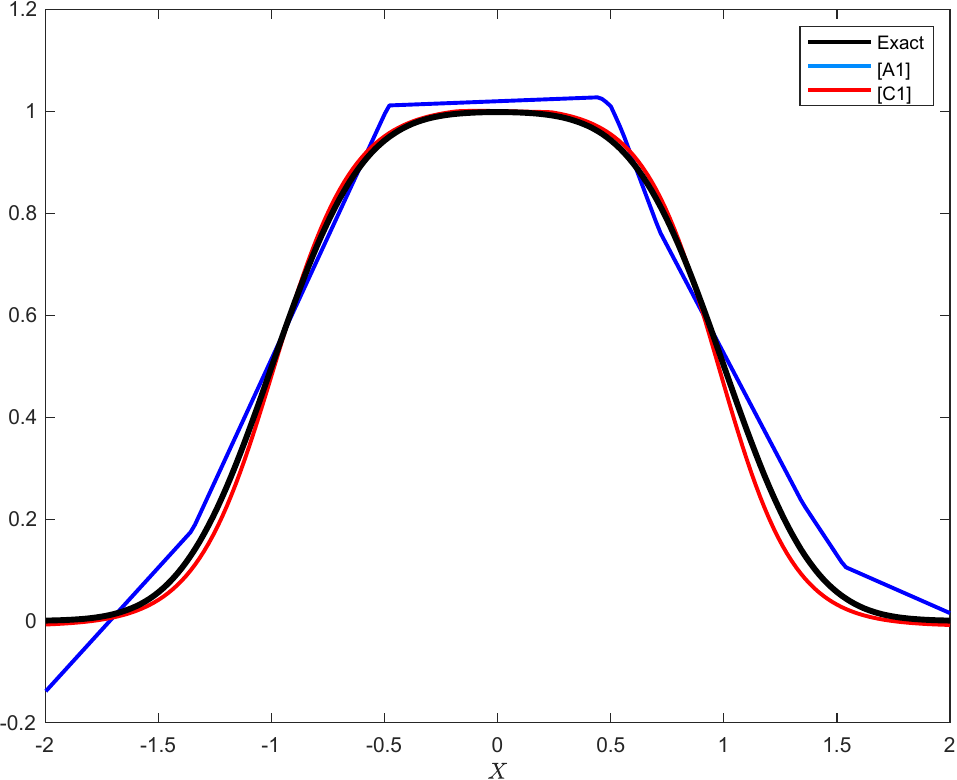}}
\centerline{(b)}
\caption{Estimation of conditional expectation $\Exp_{\cY}[\cY|\cX=X]$ when (a)~$\cY=\text{sign}(\cX)\,\cX^2+\cW$ and (b)~$\cY=\ind{[-1,1]}(\cX+\cW)$ with $n=200$ training samples and using a shallow network with hidden layer of size 50.}
\label{fig:1}
\end{figure}
To test our data-driven method we generate $n=200$ random pairs $(Y_i,X_i)$ and train a shallow neural network with a single hidden layer of size 50 and ReLU activations. To the obvious question why we select this specific size we can say that this is a fundamental problem in neural networks and currently there is no analytically trustworthy answer as to which is the appropriate network size and how must be related to the size of the training dataset. 

We apply the GD algorithm depicted in \eqref{eq:GD} and adopt the ADAM version \cite{ADAM} that normalizes each gradient element with the square root of its running power. We select a step-size equal to $\mu=0.001$ and exponential windowing with forgetting factor $\lambda=0.99$ for the power estimates. We run the algorithms for 2000 iterations which is sufficient for convergence as we could have verified if we had plotted the corresponding costs $\hat{\J}(\theta_t)$. The parameter vectors $\theta_{\mathsf{o}}$ we converge to, are used to estimate the conditional expectations as $\omega\big(\u(X,\theta_{\mathsf{o}})\big)$ and by sampling the range $[-2,2]$ of $X$ uniformly we compare with the corresponding values of the exact formulas mentioned above. 

With Example~a) since the range $\cI=\Real$, we test versions [A1], [A2], [A3] from Section\,\ref{ssec:2.A}. As we can see from Fig\,\ref{fig:1}(a) here the classical MSE version [A1] has comparable performance with the other two alternatives and all three methods approximate sufficiently well the exact funtion. Let us now turn to Example~b) where the range $\cI$ of $\cY$ and $\omega(z)$ is [0,1]. In this case we apply the classical MSE version [A1] and version [C1] with $a=-0.01,b=1.01$ which slightly overcovers $\cI$. As we observe in Fig.\,\ref{fig:1}(b) knowledge of the range and selection of the appropriate functions $\omega(z),\rho(z)$ may improve the estimation quality, dramatically.
\vskip0.1cm

\noindent\textbf{Remark~4.}~From the analytical computation of the conditional expectation it is clear that this function does not depend on the distribution of $\cX$. Since our data pairs $(Y_i,X_i)$ are sampled from the joint density, one may wonder how the marginal density of $\cX$ affects our estimates. Basically it is expected that the proposed methodology will make notable estimation errors for values of $\cX$ with small likelihood. To understand this fact, consider the extreme case where an interval of values of $\cX$ has zero probability of occurrence, then no samples from this interval can appear in the training set. It is therefore unrealistic to expect that our estimate will be accurate for such $X$ values. A similar conclusion applies when the likelihood of the interval is small and we obtain only very few (or even no) samples from the interval in the training set. This fact might seem as a serious weakness of our method however  making errors at points that \textit{never} occur or at points that occur very rarely might not be so crucial from a practical point of view.

\section{System of Equations involving Conditional Expectations}
The estimation method we introduced can be employed to compute solutions of \textit{systems of equations} expressed with the help of conditional expectations. Such systems occur in several well-known stochastic optimization problems. Characteristic examples constitute the problem of Optimal Stopping in Markov processes and the Optimal Action Policy in Reinforcement Learning which we consider in detail after introducing our general setup. 

Let $\h^j(Y,u^1,\ldots,u^K)$, $j=1,\ldots,K$ be $K$ deterministic and known scalar functions with $u^1,\ldots,u^K$ scalar variables. Suppose there are also $K$ different conditional densities $\f^j(Y|X),j=1,\ldots,K$ with $Y$ and $X$ of the \textit{same length}. Define the following system of equations which we are interested in solving for the scalar functions $\cU^1(X),\ldots,\cU^K(X)$
\begin{equation}
\cU^j(X)=\Exp_{\cY}^j\Big[\h^j\big(\cY,\cU^1(\cY),\ldots,\cU^K(\cY)\big)\big|\cX=X],
\label{eq:cond.exp.new}
\end{equation}
for $j=1,\ldots,K$ and where expectation is with respect to $\cY$ conditioned on $\cX=X$ using the conditional density $\f^j(Y|X)$.

\subsection{Numerical Solution}\label{ssec:numsol2}
We first apply the numerical method of Section\,\ref{ssec:numerical} to find a numerical solution when the $K$ conditional densities are known. We select the same sufficiently large interval for $Y$ and $X$ which we sample at the same points $Y_i=X_i,i=1,\ldots,n$. We must assure that what is left outside the interval has very small probability conditioned on every value $X_i$ and this must be true for all $K$ conditional densities. Call $\U^j=[\cU^j(Y_1),\ldots,\cU^j(Y_n)]^\intercal,j=1,\ldots,K$ the sampled version of the $K$ solution functions which we would like to determine. For each conditional density $\f^j(Y|X)$ define the matrix $\cF^j$ as explained in Section\,\ref{ssec:numerical}. Finally, form $K$ vectors $\H^j(\U^1,\ldots,\U^K), j=1,\ldots,K$ with their elements defined as follows
$$
\Big(\H^j(\U^1,\ldots,\U^K)\Big)_i=\h^j\big(Y_i^j,(\U^1)_i,\ldots,(\U^K)_i\big),
$$
$i=1,\ldots,n,~j=1,\ldots,K$,
where $(\Z)_i$ denotes the $i$th element of the vector $\Z$.
In other words the $i$th element of $\H^j$ is equal to the function $\h^j$ evaluated at $Y=Y_i^j$ with the variables $u^j$ replaced by the $i$th elements of the vectors $\U^j$. With these definitions it is clear that the sampled version of the equation in \eqref{eq:cond.exp.new} takes the following matrix/vector product form
$$
\U^j=\cF^j\times\H^j(\U^1,\ldots,\U^K),~j=1,\ldots,K.
$$
The previous system can be solved iteratively by iterating on the unknown vectors
$$
\U^j_t=\cF^j\times\H^j(\U^1_{t-1},\ldots,\U^K_{t-1}),~j=1,\ldots,K.
$$
This will be the method we are going to apply to compute the numerical solution in the problems of Optimal Stopping and Reinforcement Learning when analytic formulas are impossible.

\subsection{Data-Driven Solution}
We are now considering the data-driven version and the corresponding solution of the system of equations.
Key observation for solving the system in \eqref{eq:cond.exp.new} is that all functions of interest are defined in terms of conditional expectations and therefore, following our main idea, each conditional expectation can be estimated using the methodology we developed in Section\,\ref{ssec:ddece}.

Our data-driven setup is as follows: We are given $K$ datasets $\{(Y_1^j,X_1^j),\ldots,(Y_{n_j}^j,X_{n_j}^j)\},$ $j=1,\ldots,K$ that replace the conditional densities $\f^j(Y|X),j=1,\ldots,K$. Since we are interested in computing the functions $\cU^j(X),j=1,\ldots,K$, we approximate each function $\cU^j(X)$ with a neural network $\u(X,\theta^j)$.
Consider first the GD version for estimating the network parameters. Because of \eqref{eq:cond.exp.new} we need to apply \eqref{eq:GD} with $\d(Y)=\h_j\big(Y,\cU^1(Y),\ldots,\cU^K(Y)\big)$, namely
\begin{multline*}
\theta^j_t=\theta^j_{t-1}-\mu\sum_{i=1}^{n_j} \Big\{\h_j\big(Y^j_i,\cU^1(Y^j_i),\ldots,\cU^K(Y^j_i)\big)\allowdisplaybreaks\\ \allowdisplaybreaks-\omega\big(\u(X^j_i,\theta^j_{t-1})\big)\Big\}\rho\big(\u(X^j_i,\theta^j_{t-1})\big)\nabla_{\!\!\theta}\u(X^j_i,\theta^j_{t-1}). \allowdisplaybreaks
\end{multline*}
Unfortunately the previous formula is impossible to use for the parameter updates because the functions $\cU^j(Y)$ on the right hand side are the ones we are actually attempting to estimate. We recall that our estimation method of conditional expectation results in a final approximation of the form $\omega\big(\u(X,\theta^j_{\mathsf{o}})\big)\approx\cU^j(X)$ consequently, due to \eqref{eq:cond.exp.new}, $\omega\big(\u(Y,\theta^j_{\mathsf{o}})\big)$ could replace $\cU^j(Y)$ in the previous updates. But this is still problematic since the limits $\theta^j_{\mathsf{o}}$ are not known in advance. We therefore propose at iteration $t$ to employ the most recent estimate $\omega\big(\u(Y,\theta^j_{t-1})\big)$ of $\cU^j(Y)$. This selection clearly allows for computations and produces the following updates
\begin{multline}
\theta^j_t=\theta^j_{t-1}-\\
\mu\sum_{i=1}^{n_j} \Big\{\h_j\Big(Y^j_i,\omega\big(\u(Y^j_i,\theta^1_{t-1})\big),\ldots,\omega\big(\u(Y_i^j,\theta^K_{t-1})\big)\Big)\\
-\omega\big(\u(X^j_i,\theta^j_{t-1})\big)\Big\}
\rho\big(\u(X^j_i,\theta^j_{t-1})\big)\nabla_{\!\!\theta}\u(X^j_i,\theta^j_{t-1}),
\label{eq:dd-gd}
\end{multline}
for $j=1,\ldots,K$.
We note that in the iteration for $\theta^j_t$ we use the dataset corresponding to $\f^j(Y|X)$. We also observe that the updates are performed in parallel since each iteration involves the update of all $K$ network parameter vectors $\theta^j_t,j=1,\ldots,K$. Of course it is not necessary to use the same $\omega(\cdot),\rho(\cdot)$ functions or the same neural network configuration when approximating the desired functions $\cU^j(X)$.

In case we prefer to employ the SGD then the $K$ datasets must be (randomly) mixed with each pair retaining its label. Then at iteration $t$ if we select to process the pair $(Y_t,X_t)$ with label $j$ we apply
\begin{align*}
\theta^j_t&=\theta^j_{t-1}-\mu\Big\{\h_j\Big(Y_t,\omega\big(\u(Y_t,\theta^1_{t-1})\big),\ldots,\omega\big(\u(Y_t,\theta^K_{t-1})\big)\Big)\\
&~~~~~-\omega\big(\u(X_t,\theta^j_{t-1})\big)\Big\}
\rho\big(\u(X_t,\theta^j_{t-1})\big)\nabla_{\!\!\theta}\u(X_t,\theta^j_{t-1})\\[2pt]
\theta^\ell_t&=\theta^\ell_{t-1},\ell\neq j.
\end{align*}
The limiting values $\theta^j_{\mathsf{o}},~j=1,\ldots,K$ provide the estimates $\omega\big(\u(X,\theta^j_{\mathsf{o}})\big)\approx\cU^j(X)$ which are approximations of the solution of the system of equations.

This is the general data-driven approach and computational methodology we propose. To test its effectiveness, we apply it to Optimal Stopping and Reinforcement Learning.

\subsection{Markov Optimal Stopping}\label{ssec:MOS}
Consider a homogeneous Markov process $\{\cX_t\}$ which is observed (sampled) sequentially, namely at each time $t$ we observe a new point $\cX_t$. We can decide to stop sampling at any time $T$ which can either be deterministic or random. When the decision to stop at $\{T=t\}$ is based on the available information $\{X_0,\ldots,X_t\}$ accumulated up to time $t$, then $T$ is called a \textit{stopping time adapted to} $\{\cX_t\}$.  

We are interested in the minimization of the following exponentially discounted average cost over a stopping time~$T$ \cite{PS,ANS}
$$
\cU(X)=\inf_T\Exp\Big[\alpha^T\p(\cX_T)+\sum_{t=0}^{T-1}\alpha^t\q(\cX_t)\Big|X_0=X\Big],
$$
where $\alpha$ is the exponential factor, $\p(\cdot),\q(\cdot)$ are known functions with $\q(\cX_t)$ expressing the cost of sampling at $t$ and $\p(\cX_t)$ the cost of stopping at $t$. If we consider the infinite horizon version (no hard limit on $T$) then we know (see \cite{TNB}, Page 70) that $\cU(X)$ satisfies the \textit{equation}
\begin{equation}
\cU(X)=\Exp_{\cX_1}\big[\min\big\{\p(\cX_1),\q(\cX_1)+\alpha\cU(\cX_1)\big\}|X_0=X\big],
\label{eq:optstop}
\end{equation}
while the stopping time $T_{\mathsf{o}}$ defined as
\begin{equation}
T_{\mathsf{o}}=\inf\big\{t\geq0:\p(X_t)\leq\q(X_t)+\alpha\cU(X_t)\big\},
\label{eq:stoptime}
\end{equation}
delivers the optimal cost $\cU(X)$ and is therefore optimum. The general system in \eqref{eq:cond.exp.new} clearly covers equation \eqref{eq:optstop} for Optimal Stopping by selecting $K=1$ and $\h^1(Y,u^1)=\min\{\p(Y),\q(Y)+\alpha u^1\}$. 

A numerical solution for this problem can be devised based on the method introduced in Section\,\ref{ssec:numsol2} when the transition density is known and $X$ is scalar. If we select a sufficiently large interval and sample it at $\{X_1,\ldots,X_n\}$ we generate the vector $\U=[\cU(X_1),\ldots,\cU(X_n)]^\intercal$ of samples from the unknown solution, the known vectors $\P=[\p(X_1),\ldots,\p(X_n)]^\intercal$,
$\Q=[\q(X_1),\ldots,\q(X_n)]^\intercal$ and, finally the matrix $\cF$ as detailed in Section\,\ref{ssec:numerical}. Then the sampled version of \eqref{eq:optstop} takes the form
$$
\U=\cF\times\min\{\P,\Q+\alpha\U\}
$$
where the ``min'' is applied on an element-by-element basis on the two vectors $\P$ and $\Q+\alpha\U$. This equation can be solved by iterating over $\U$ as follows
\begin{equation}
\U_t=\cF\times\min\{\P,\Q+\alpha\U_{t-1}\},~~\U_0=\P,
\label{eq:os_num}
\end{equation}
and considering the limit of $\U_t$ as approximating the sampled form of the desired solution.

\subsubsection*{\underline{Example with AR(1) Process}}
\begin{figure}[t!]
\centerline{\includegraphics[scale=0.48]{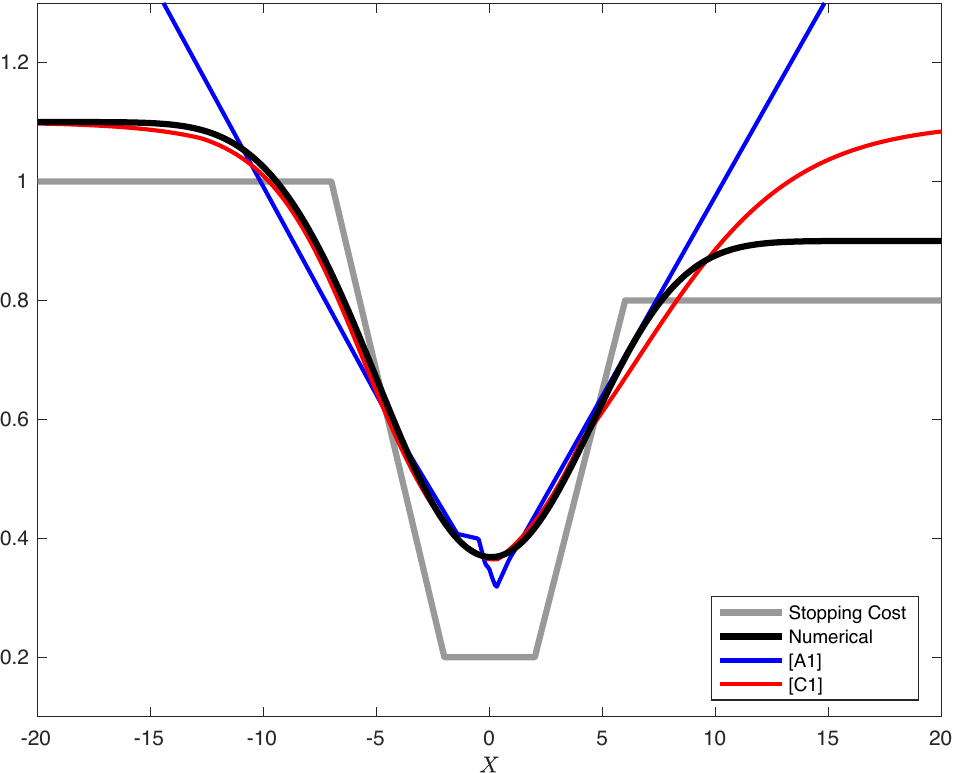}}
\centerline{(a)}
\vskip0.4cm
\centerline{\includegraphics[scale=0.48]{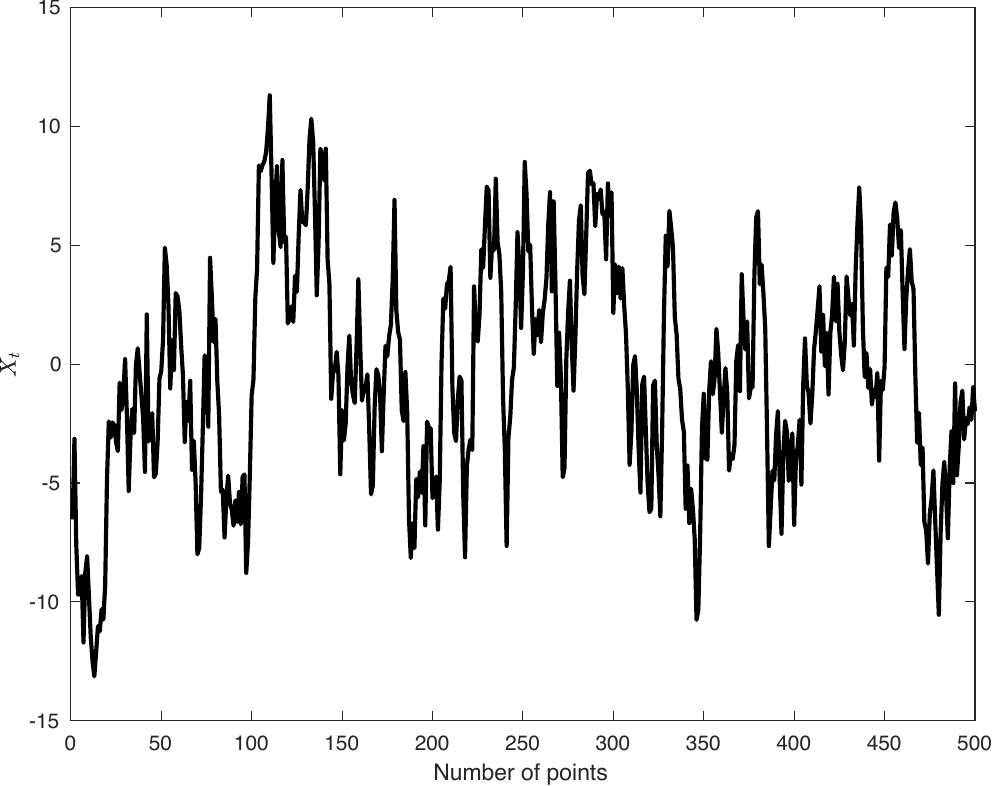}}
\centerline{(b)}
\caption{(a)~Numerical and Data-Driven estimation of the optimal average cost with $n=500$ samples, using a shallow networks with hidden layer of size 100 nodes. (b)~The training set with 500 consecutive realization of the Markov process $\{\cX_t\}$.}
\label{fig:2}
\end{figure}

Let $\{\cX_t\}$ be a homogeneous AR(1) process of the form $\cX_t=0.9\,\cX_{t-1}+\sqrt{5}\cW_t$ where $\{\cW_t\}$ is standard i.i.d.~Gaussian noise sequence. The sampling cost is selected $\q(X)=0.1$ while the stopping cost $\p(X)$ is depicted in Fig.\,\ref{fig:2}(a) in gray. The exponential discount factor is set to $\alpha=1$. Because of this selection we can prove that $\min_X\p(X)\leq\cU(X)\leq\max_X\p(X)$ which can define a possible range $\cI$ for the solution.

Since solving \eqref{eq:optstop} analytically is impossible we apply the numerical method we detailed above. In particular we select the interval [-30,30] and sample it uniformly at 5000 points. We form the matrix $\cF$ and the vectors $\U,\P,\Q$ and iterate over $\U_t$ for 1000 times. In Fig.\,\ref{fig:2}(a), as we said, we can see the stopping cost $\p(X)$ in gray and the numerical solution elevated by the sampling cost $\q(X)$ in black. In other words we plot $\Q+\U$ because, according to \eqref{eq:stoptime}, this sum must be compared with the stopping cost in order to decide optimally whether to stop or continue sampling at any time $t$.

For our proposed data-driven estimation method, we generate 500 consecutive realizations of the AR(1) process and use them to train a shallow network with hidden layer of size 100 and ReLU activations. We apply the GD algorithm depicted in \eqref{eq:dd-gd} with $K=1$, step-size $\mu=0.001$ and forgetting factor for the ADAM version equal to $\lambda=0.99$. We run [A1] and [C1] with $\cI=[0.2,1]$ (the lower and upper bound of $\p(X)$) for 2000 iterations. Version [A1] (MSE) is plotted with blue line whereas [C1] with red (both elevated by $\q(X)$).
We realize again that knowledge of the range can result in better estimates. Observing Fig.\,\ref{fig:2}(a) one may argue that the error in [C1] is more pronounced when $X\geq10$. This is because, as we can see from Fig.\,\ref{fig:2}(b) where we plot the 500 consecutive realizations of the Markov process, there are very few samples in the training set with such values. According to Remark\,4, this is the reason why we may experience estimates with large errors.

\subsection{Optimal Action Policy in Reinforcement Learning}
A second perhaps more popular nowadays problem is the optimal action policy in Reinforcement Learning. Suppose $\{\cS_t\}$ is a Markov controlled process with the action taking discrete values in the set $\{1,2,\ldots,K\}$ and with each action corresponding to a different transition density. Denote with $\Exp^j[\cdot]$ the conditional expectation with respect to the transition density of action $j$. With every state $S$ assume there is a reward $\cR(S)$ where $\cR(\cdot)$ is a known scalar deterministic function. We are looking for the best action policy that will result in maximal average reward over an infinite and exponentially discounted time horizon of the form
\begin{align*}
\cQ(S)&=\max_{j_0,j_1,\ldots}\Exp[\cR(\cS_1)+\gamma\cR(\cS_2)+\gamma^2\cR(\cS_3)+\cdots\}|\cS_0=S]\\
&=\max_{j_0}\Exp_{\cS_1}[\cR(\cS_1)+\gamma\cQ(\cS_1)|\cS_0=S]
\end{align*}
where $0\leq\gamma<1$ is the geometric discount factor. 

If at time $t=0$ we observe state $\cS_0=S$ and we decide in favor of action $j$ and \textit{after this point we always use optimal action policy} then let us call the resulting reward $\cU^j(S)$. It can then be proved (see \cite{SB}, Eq.~(4.2), Page 90) 
that these functions satisfy the following \textit{system of equations}
\begin{equation}
\cU^j(S)=\Exp_{\cS_1}^j\Big[\cR(\cS_1)+\gamma
\max_{1\leq\ell\leq K}\cU^\ell(\cS_1)\big|\cS_0=S\Big],
\label{eq:RL3}
\end{equation}
where $j=1,\ldots,K$. Also
\begin{equation}
\cQ(S)=\max_{1\leq \ell\leq K}\cU^\ell(S).
\label{eq:RL2}
\end{equation}
The previous equality suggests that if we know the functions $\cU^j(S)$, then at time $t$ if we observe state $\cS_t=S_t$ the action $j_t^{\mathsf{o}}$ which guarantees optimal reward is
$$
j_t^{\mathsf{o}}=\text{arg}\max_{1\leq\ell\leq K}\cU^\ell(S_t).
$$
The system of equations \eqref{eq:RL3} is a special case of the general system in \eqref{eq:cond.exp.new}. Indeed, we need to select all functions to be the same, that is, $\h^j(\cdot)=\h(\cdot)$ with $\h(Y,u^1,\ldots,u^K)=\cR(Y)+\gamma\max_{1\leq \ell\leq K}u^\ell$.

It is only for simplicity we have considered a finite number of actions. Applying similar analysis we can accommodate continuous actions with the corresponding optimal rewards $\cU^j(S)$ replaced by the function $\cU(S,a)$ and $a$ denoting action with continuous value.

As in the Markov optimal stopping problem, we can offer a numerical solution when the $K$ transition densities $\f^j(S_t|S_{t-1})$ are known and the state is scalar. Again we select a sufficiently large interval which we sample at the points $\{S_1,\ldots,S_n\}$. This gives rise to the $K$ vectors $\U^j=[\cU^j(S_1),\ldots,\cU^j(S_n)]^\intercal$ which are the sampled form of the $K$ solution functions. We also form the $K$ matrices $\cF^j$ using the corresponding transition densities as explained in Section\,\ref{ssec:numerical}. Finally we consider the single reward vector $\R=[\cR(S_1),\ldots,\cR(S_n)]^\intercal$. The system of equations \eqref{eq:RL3} under a sampled form becomes
$$
\U^j=\cF^j\times\big\{\R+\gamma\max_{1\leq\ell\leq K}\U^{\ell}\big\},~j=1,\ldots,K,
$$
where the ``max'' is taken on an element-by-element basis over the $K$ vectors $\U^\ell$. The solution of this system of equations can be obtained by applying the following iterative scheme
\begin{equation}
\U^j_t=\cF^j\times\big\{\R+\gamma\max_{1\leq\ell\leq K}\U_{t-1}^{\ell}\big\},~~\U_0^j=0,
\label{eq:rl_num}
\end{equation}
where $j=1,\ldots,K$, with the limits (as $t\to\infty$) of the $K$ vector sequences $\{\U_t^j\}, j=1,\ldots,K$ approximating the sampled version of the optimal functions $\cU^j(S)$. Let us now apply this methodology to a specific example.

\subsubsection*{\underline{Example with Two AR(1) Processes}}
We present a simple example with $K=2$ actions. The corresponding Markov processes are both AR(1) and of the form 1)~$\cS_t=0.8\cS_{t-1}+1+\cW_t$, 2)~$\cS_t=0.8\cS_{t-1}-1+\cW_t$ with $\{\cW_t\}$ i.i.d.~standard Gaussians. We consider a reward function $\cR(S)$ which is the same as the stopping cost of the previous example and depicted in Fig.\,\ref{fig:2}(a) in gray. We also set the geometric discount factor equal to $\gamma=0.8$. 

\begin{figure}[b!]
\centerline{\includegraphics[scale=0.48]{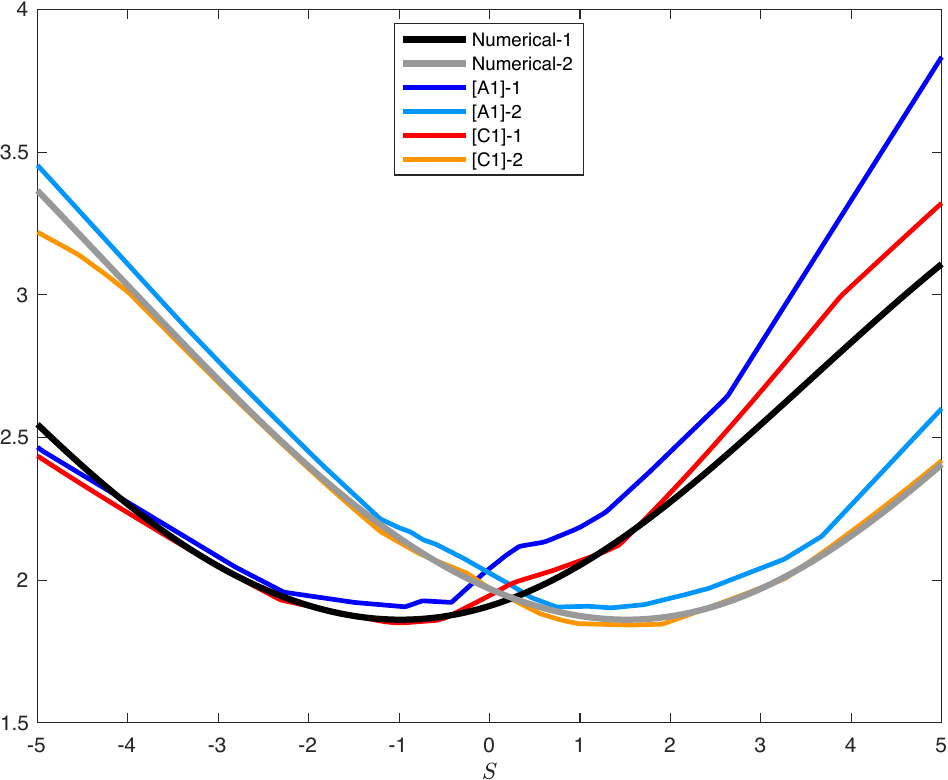}}
\centerline{(a)}
\vskip0.4cm
\centerline{\includegraphics[scale=0.48]{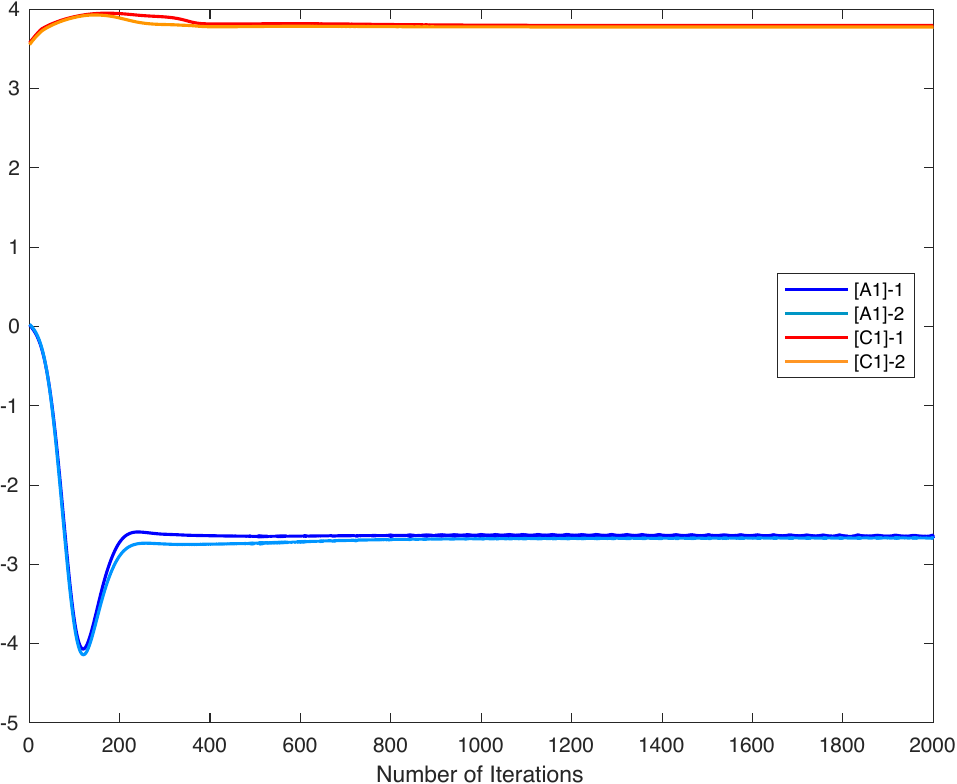}}
\centerline{(b)}
\caption{(a)~Numerical and Data-Driven estimation of the two expected optimal reward function $\cU^1(S),\cU^2(S)$ with $n=500$ average number of samples per model, using two shallow networks with hidden layer of size 100 nodes. (b)~Evolution of costs as a function of the number of iterations.}
\label{fig:3}
\end{figure}
For the numerical solution we select the interval $[-20,20]$ which we sample at 5000 points and apply the iteration presented in \eqref{eq:rl_num}. The outcome can be seen in Fig.\,\ref{fig:3}(a) with black and gray lines for $\cU^1(S),\cU^2(S)$ respectively. It is against these results that we need to compare the data-driven method.

For the data-driven estimation we randomly generate a length $n=1000$ action sequence $\{a_t\}$ with $a_t\in\{1,2\}$ which is used to generate 1000 realizations for the Markov controlled process. If $a_t=i$ then to go from $S_t$ to $S_{t+1}$ we use the $i$th Markov model. This means that on average we have 500 points per model. We apply [A1] and [C1] and for the latter, since $0.2\leq\cR(S)\leq1$, we can show that $\frac{0.2}{1-\gamma}\leq\cU^j(X)\leq\frac{1}{1-\gamma}$ suggesting that we can select the interval $\cI=[1,5]$ as our range. For both functions we select a shallow neural network with hidden layer of size 100 and ReLU activations. We run the algorithms for 2000 iterations using a step size $\mu=0.001$ and forgetting factor for the power estimation $\lambda=0.99$ of the ADAM version. Again knowledge of the range and using it in [C1] can produce better estimates. From Fig.\,\ref{fig:3}(b) by following the evolution of the costs, we conclude that both iteration methods are stable and converge successfully.

\subsubsection*{\underline{Exploration vs Exploitation}}
Reinforcement Learning has become very popular due to the possibility of \textit{Exploring} and \textit{Exploiting}.
Assuming that we start with sufficient number of initial data from each action, we can use them as described above to make an initial estimate of the functions $\cU^j(S)$. Then these estimates can be used to decide about the next actions which constitutes the \textit{Exploitation} phase. 

If the initial data are not sufficient to guarantee an estimate of acceptable accuracy, or if the statistical behavior of the data changes with time then we must periodically make \textit{random} decisions about the next action in order to activate possibilities that are not reachable by the existing action rule. These randomly generated data must be used to update the estimates of the functions $\cU^j(S)$ and this is achieved mostly with the help of the SGD. This is known as the \textit{Exploration} phase.

Whether at each step we should explore or exploit can be decided using randomization. Specifically, at each time $t$ with probability $1-\epsilon$ we could exploit and with probability $\epsilon$ explore. An asymptotic analysis could provide the limiting behavior of a scheme of this form in the stationary case when there is no change in the statistical behavior of the data but also in the case where the data are (slowly) varying in time and the updates attempt to \textit{track} the change. Such an analysis is already available for adaptive algorithms for classical FIR filters \cite{EF,NGW} and it would be extremely interesting if we were able to extend it to accommodate the neural network class.



\appendices
\onecolumn
\section{Matlab Code for Conditional Expectation Examples}
\noindent\normalsize Copy/Paste from the PDF file to the Matlab (or text) editor needs extra work. The symbols `` $^\wedge$ '' and `` $^\prime$ '' used for power and transpose are transferred wrongly. In the editor you need to apply a global replace with the corresponding keyboard symbols.
\vskip0.1cm
\noindent Run \texttt{ce\textunderscore main1.m} and \texttt{ce\textunderscore main2.m} for the first and second example appearing in Fig.\,\ref{fig:1}(a) and (b).
\vskip0.1cm
\noindent\scriptsize\noindent\textbf{ce\textunderscore main1.m}\\[-13pt]
\begin{verbatim}
n=200;
L=50;
times=2000;
X=randn(n,1); % Input
s=0.1; % Noise variance
W=sqrt(s)*randn(n,1); % Noise
Y=sign(X).*((X).^+W; % Noisy data
% Exact Conditional Expectation
Xo=-2+4*[0:200]'/200;
Yo=sign(Xo).*((Xo).^2);
figure(1)
plot(Xo,Yo,'k','linewidth',3); % Exact
ce_a1  % [A1]
Z1=A0*Xo'+B0;
X1=max(Z1,0);
Z=A1'*X1+B1;
Ye=Z;
figure(1)
hold on
plot(Xo,Ye,'b','linewidth',2);
hold off
ce_a2  % [A2]
Z1=A0*Xo'+B0;
X1=max(Z1,0);
Z=A1'*X1+B1;
Ye=sinh(Z);
figure(1)
hold on
plot(Xo,Ye,'r','linewidth',2);
hold off
ce_a3  % [A3]
Z1=A0*Xo'+B0;
X1=max(Z1,0);
Z=A1'*X1+B1;
Ye=sign(Z).*(exp(abs(Z))-1);
figure(1)
hold on
plot(Xo,Ye,'g','linewidth',2);
hold off
xlabel('$X$','interpreter','latex')
legend('Exact','[A1]','[A2]','[A3]','location','southeast')
\end{verbatim}

\scriptsize\noindent\textbf{ce\textunderscore main2.m}\\[-13pt]
\begin{verbatim}
n=200;
L=50;
times=2000;
X=randn(n,1); % Input
s=0.1; % Noise variance
W=sqrt(s)*randn(n,1); % Noise
Y=double([X+W>=-1])-double([X+W>1]); % Noisy data
% Exact Conditional Expectation
Xo=-2+4*[0:200]'/200;
Yo=normcdf((1-Xo)/sqrt(s))-normcdf((-1-Xo)/sqrt(s));
figure(1)
plot(Xo,Yo,'k','linewidth',3); % Exact
ce_a1  % [A1]
Z1=A0*Xo'+B0;
X1=max(Z1,0);
Z=A1'*X1+B1;
Ye=Z;
figure(1)
hold on
plot(Xo,Ye,'b','linewidth',2);
hold off
a=-0.01;
b=1.01;
ce_c1  % [C1]
Z1=A0*Xo'+B0;
X1=max(Z1,0);
Z=A1'*X1+B1;
Ye=a./(1+exp(Z))+b./(1+exp(-Z));
figure(1)
hold on
plot(Xo,Ye,'r','linewidth',2);
hold off
xlabel('$X$','interpreter','latex')
legend('Exact','[A1]','[C1]','location','northeast')
\end{verbatim}

\scriptsize\noindent\textbf{ce\textunderscore a1.m}\\[-13pt]
\begin{verbatim}
A0=randn(L,1)/sqrt(L);  % Network parameters
B0=zeros(L,1);
A1=randn(L,1)/sqrt(L);
B1=0;
PDA0=zeros(L,1); % Parameter powers
PDB0=zeros(L,1);
PDA1=zeros(L,1);
PDB1=0;
mu=0.001; % Step size
la=0.99; % Forgetting factor ADAM
c=0.001; % To avoid division by 0
cost=zeros(times,1);
for ttt=1:times
    Z1=A0*X'+B0;
    X1=max(Z1,0);
    Z2=A1'*X1+B1;
    cost(ttt)=mean(Z2.^2/2-Y'.*Z2);
    % Computations of derivatives
    U=-(Y-Z2');
    DA1=X1*U;
    DB1=sum(U);
    DB0=A1.*(max(sign(Z1),0)*U);
    DA0=A1.*((max(sign(Z1),0).*X')*U);
    % Estimation of Powers for ADAM
    if ttt==1
        PDA1=DA1.^2;
        PDB1=DB1.^2;
        PDA0=DA0.^2;
        PDB0=DB0.^2;
    else
        PDA1=la*PDA1+(1-la)*DA1.^2;
        PDB1=la*PDB1+(1-la)*DB1.^2;
        PDA0=la*PDA0+(1-la)*DA0.^2;
        PDB0=la*PDB0+(1-la)*DB0.^2;
    end
    A0=A0-mu*(DA0./sqrt(c+PDA0));
    B0=B0-mu*(DB0./sqrt(c+PDB0));
    A1=A1-mu*(DA1./sqrt(c+PDA1));
    B1=B1-mu*(DB1./sqrt(c+PDB1));
end
\end{verbatim}

\scriptsize\noindent\textbf{ce\textunderscore a2.m}\\[-13pt]
\begin{verbatim}
A0=randn(L,1)/sqrt(L);  % Network parameters
B0=zeros(L,1);
A1=randn(L,1)/sqrt(L);
B1=0;
PDA0=zeros(L,1); % Parameter powers
PDB0=zeros(L,1);
PDA1=zeros(L,1);
PDB1=0;
mu=0.001; % Step size
la=0.99; % Forgetting factor ADAM
c=0.001; % To avoid division by 0
cost=zeros(times,1);
for ttt=1:times
    Z1=A0*X'+B0;
    X1=max(Z1,0);
    Z2=A1'*X1+B1;
    cost(ttt)=mean(exp(abs(Z2)/2)-1+(exp(-1.5*abs(Z2))-1)/3+2*Y'.*sign(Z2).*(exp(-abs(Z2)/2)-1));
    % Computations of derivatives
    U=-exp(-abs(Z2')/2).*(Y-sinh(Z2'));
    DA1=X1*U;
    DB1=sum(U);
    DB0=(A1.*max(sign(Z1),0))*U;
    DA0=(A1.*max(sign(Z1),0))*(X.*U);
    % Estimation of Powers for ADAM
    if ttt==1
        PDA1=DA1.^2;
        PDB1=DB1.^2;
        PDA0=DA0.^2;
        PDB0=DB0.^2;
    else
        PDA1=la*PDA1+(1-la)*DA1.^2;
        PDB1=la*PDB1+(1-la)*DB1.^2;
        PDA0=la*PDA0+(1-la)*DA0.^2;
        PDB0=la*PDB0+(1-la)*DB0.^2;
    end
    A0=A0-mu*(DA0./sqrt(c+PDA0));
    B0=B0-mu*(DB0./sqrt(c+PDB0));
    A1=A1-mu*(DA1./sqrt(c+PDA1));
    B1=B1-mu*(DB1./sqrt(c+PDB1));
end
\end{verbatim}
\newpage
\scriptsize\noindent\textbf{ce\textunderscore a3.m}\\[-13pt]
\begin{verbatim}
A0=randn(L,1)/sqrt(L);  % Network parameters
B0=zeros(L,1);
A1=randn(L,1)/sqrt(L);
B1=0;
PDA0=zeros(L,1); % Parameter powers
PDB0=zeros(L,1);
PDA1=zeros(L,1);
PDB1=0;
mu=0.001; % Step size
la=0.99; % Forgetting factor ADAM
c=0.001; % To avoid division by 0
cost=zeros(times,1);
for ttt=1:times
    Z1=A0*X'+B0;
    X1=max(Z1,0);
    Z2=A1'*X1+B1;
    cost(ttt)=mean(4*cosh(Z2/2)+2*Y'.*sign(Z2).*(exp(-abs(Z2)/2)-1));
    % Computations of derivatives
    U=-exp(-abs(Z2')/2).*(Y-sign(Z2').*(exp(abs(Z2'))-1));
    DA1=X1*U;
    DB1=sum(U);
    DB0=(A1.*max(sign(Z1),0))*U;
    DA0=(A1.*max(sign(Z1),0))*(X.*U);
    % Estimation of Powers for ADAM
    if ttt==1
        PDA1=DA1.^2;
        PDB1=DB1.^2;
        PDA0=DA0.^2;
        PDB0=DB0.^2;
    else
        PDA1=la*PDA1+(1-la)*DA1.^2;
        PDB1=la*PDB1+(1-la)*DB1.^2;
        PDA0=la*PDA0+(1-la)*DA0.^2;
        PDB0=la*PDB0+(1-la)*DB0.^2;
    end
    A0=A0-mu*(DA0./sqrt(c+PDA0));
    B0=B0-mu*(DB0./sqrt(c+PDB0));
    A1=A1-mu*(DA1./sqrt(c+PDA1));
    B1=B1-mu*(DB1./sqrt(c+PDB1));
end
\end{verbatim}

\scriptsize\noindent\textbf{ce\textunderscore c1.m}\\[-13pt]
\begin{verbatim}
A0=randn(L,1)/sqrt(L);  % Network parameters
B0=zeros(L,1);
A1=randn(L,1)/sqrt(L);
B1=0;
PDA0=zeros(L,1); % Parameter powers
PDB0=zeros(L,1);
PDA1=zeros(L,1);
PDB1=0;
mu=0.001; % Step size
la=0.99; % Forgetting factor ADAM
c=0.001; % To avoid division by 0
cost=zeros(times,1);
for ttt=1:times
    Z1=A0*X'+B0;
    X1=max(Z1,0);
    Z2=A1'*X1+B1;
    cost(ttt)=mean((b-a)./(1+exp(Z2))+b*log(1+exp(Z2))-Y'.*log(1+exp(Z2)));
    % Computations of derivatives
    U=-(1./(1+exp(-Z2'))).*(Y-b./(1+exp(-Z2'))-a./(1+exp(Z2')));
    DA1=X1*U;
    DB1=sum(U);
    DB0=(A1.*max(sign(Z1),0))*U;
    DA0=(A1.*max(sign(Z1),0))*(X.*U);
    % Estimation of Powers for ADAM
    if ttt==1
        PDA1=DA1.^2;
        PDB1=DB1.^2;
        PDA0=DA0.^2;
        PDB0=DB0.^2;
    else
        PDA1=la*PDA1+(1-la)*DA1.^2;
        PDB1=la*PDB1+(1-la)*DB1.^2;
        PDA0=la*PDA0+(1-la)*DA0.^2;
        PDB0=la*PDB0+(1-la)*DB0.^2;
    end
    A0=A0-mu*(DA0./sqrt(c+PDA0));
    B0=B0-mu*(DB0./sqrt(c+PDB0));
    A1=A1-mu*(DA1./sqrt(c+PDA1));
    B1=B1-mu*(DB1./sqrt(c+PDB1));
end
\end{verbatim}
\vfill
\newpage

\section{Matlab Code for Optimal Stopping Example}
\noindent\normalsize Copy/Paste from the PDF file to the Matlab (or text) editor needs extra work. The symbols `` $^\wedge$ '' and `` $^\prime$ '' used for power and transpose are transferred wrongly. In the editor you need to apply a global replace with the corresponding keyboard symbols.
\vskip0.1cm
\normalsize\noindent Run \texttt{os\textunderscore main.m} for the example appearing in Fig.\,\ref{fig:2}(a) and (b).
\vskip0.1cm
\noindent\scriptsize\noindent\textbf{os\textunderscore main.m}\\[-13pt]
\begin{verbatim}
r=0.9;
s=5.0;
m=0.0;
a=0.2;
b=1.0;
mu=0.001;
la=0.99;
c=0.001;
n=500;
L=100;
times=2000;
os_num;  % Numerical solution
cX=zeros(n+1,1); % Markov process
cX(1)=randn*sqrt(s)/sqrt(1-r^2);
for iii=2:n+1
    cX(iii)=r*cX(iii-1)+sqrt(s)*randn+m;
end
X=cX(1:end-1);
Y=cX(2:end);
pY=stopcost(Y);
qY=sampcost(Y);
os_a1   %[A1]
XX=-20+40*[0:500]'/500; 
qXX=sampcost(XX);
Z1=A0*XX'+B0;
X1=max(Z1,0);
Z2=A1'*X1+B1;
figure(1)
hold on
plot(XX,qXX+Z2','b','linewidth',2)
hold off
os_c1   %[C1]
Z1=A0*XX'+B0;
X1=max(Z1,0);
Z2=A1'*X1+B1;
Z2o=a./(1+exp(Z2'))+b./(1+exp(-Z2'));
figure(1)
hold on
plot(XX,qXX+Z2o,'r','linewidth',2)
hold off
ax=axis;
My=max(max(qXX+Z2o),1.2);
my=min(min(qXX+Z2o),0.2);
ax(3:4)=[my-0.1 My+0.1];
axis(ax)
legend('Stopping Cost','Numerical','[A1]','[C1]')
figure(2);plot([cost_a1 cost_c1]);
\end{verbatim}

\noindent\scriptsize\noindent\textbf{os\textunderscore num.m}\\[-13pt]
\begin{verbatim}
ntimes=1000;
X=-30+60*[0:5000]'/5000;
F=normcdf((X'-r*X-m)/sqrt(s));
[max(F(:,1)) max(1-F(:,end))]
F(:,1)=zeros(length(X),1);
F(:,end)=ones(length(X),1);
F=[F(:,2)-F(:,1) F(:,3:end)-F(:,1:end-2) F(:,end)-F(:,end-1)]/2;
U=stopcost(X);
q=sampcost(X);
p=stopcost(X);
for iii=1:ntimes
    U=F*min(p,q+U);
end
I=[X>=-20]&[X<=20];
figure(1);
plot(X(I),p(I),'color',[0.6 0.6 0.6],'linewidth',3);
hold on
plot(X(I),q(I)+U(I),'k','linewidth',3);
hold off
\end{verbatim}

\noindent\scriptsize\noindent\textbf{os\textunderscore a1.m}\\[-13pt]
\begin{verbatim}
A0=randn(L,1)/sqrt(L);
B0=zeros(L,1);
A1=randn(L,1)/sqrt(L);
B1=0;
PDA0=zeros(L,1);
PDB0=zeros(L,1);
PDA1=zeros(L,1);
PDB1=0;
cost_a1=zeros(times,1);
for ttt=1:times
    Z1=A0*X'+B0;
    X1=max(Z1,0);
    Z2=A1'*X1+B1;
    ZY1=A0*Y'+B0;
    XY1=max(ZY1,0);
    ZY2=A1'*XY1+B1;
    YY=min(pY,qY+ZY2');
    cost_a1(ttt)=mean(Z2.^2/2-YY'.*Z2);
    % Computations of derivatives
    U=-(YY-Z2');
    DA1=X1*U;
    DB1=sum(U);
    DB0=A1.*(max(sign(Z1),0)*U);
    DA0=A1.*((max(sign(Z1),0).*X')*U);
    % Power updates
    if ttt==1
        PDA1=DA1.^2;
        PDB1=DB1.^2;
        PDA0=DA0.^2;
        PDB0=DB0.^2;
    else
        PDA1=la*PDA1+(1-la)*DA1.^2;
        PDB1=la*PDB1+(1-la)*DB1.^2;
        PDA0=la*PDA0+(1-la)*DA0.^2;
        PDB0=la*PDB0+(1-la)*DB0.^2;
    end
    % Parameter updates
    A0=A0-mu*(DA0./sqrt(c+PDA0));
    B0=B0-mu*(DB0./sqrt(c+PDB0));
    A1=A1-mu*(DA1./sqrt(c+PDA1));
    B1=B1-mu*(DB1./sqrt(c+PDB1));
end
\end{verbatim}

\noindent\scriptsize\noindent\textbf{os\textunderscore c1.m}\\[-13pt]
\begin{verbatim}
A0=randn(L,1)/sqrt(L);
B0=zeros(L,1);
A1=randn(L,1)/sqrt(L);
B1=0;
PDA0=zeros(L,1);
PDB0=zeros(L,1);
PDA1=zeros(L,1);
PDB1=0;
cost_c1=zeros(times,1);
for ttt=1:times
    Z1=A0*X'+B0;
    X1=max(Z1,0);
    Z2=A1'*X1+B1;
    om_Z2=a./(1+exp(Z2'))+b./(1+exp(-Z2'));
    ZY1=A0*Y'+B0;
    XY1=max(ZY1,0);
    ZY2=A1'*XY1+B1;
    om_ZY2=a./(1+exp(ZY2'))+b./(1+exp(-ZY2'));
    YY=min(pY,qY+om_ZY2);
    cost_c1(ttt)=mean((b-a)./(1+exp(Z2))+b*log(1+exp(Z2))-YY'.*log(1+exp(Z2)));
    % Computations of derivatives
    U=-(1./(1+exp(-Z2'))).*(YY-om_Z2);
    DA1=X1*U;
    DB1=sum(U);
    DB0=(A1.*max(sign(Z1),0))*U;
    DA0=(A1.*max(sign(Z1),0))*(X.*U);
    if ttt==1
        PDA1=DA1.^2;
        PDB1=DB1.^2;
        PDA0=DA0.^2;
        PDB0=DB0.^2;
    else
        PDA1=la*PDA1+(1-la)*DA1.^2;
        PDB1=la*PDB1+(1-la)*DB1.^2;
        PDA0=la*PDA0+(1-la)*DA0.^2;
        PDB0=la*PDB0+(1-la)*DB0.^2;
    end
    A0=A0-mu*(DA0./sqrt(c+PDA0));
    B0=B0-mu*(DB0./sqrt(c+PDB0));
    A1=A1-mu*(DA1./sqrt(c+PDA1));
    B1=B1-mu*(DB1./sqrt(c+PDB1));
end
\end{verbatim}

\noindent\scriptsize\noindent\textbf{stopcost.m}\\[-13pt]
\begin{verbatim}
function pX=stopcost(X);
pX=zeros(length(X),1);
for i=1:length(X)
    if X(i)<-7;pX(i)=1;end
    if (X(i)>=-7)&(X(i)<-2);pX(i)=1-(X(i)+7)*0.8/5;end
    if (X(i)>=-2)&(X(i)<=2);pX(i)=0.2;end
    if (X(i)>2)&(X(i)<6);pX(i)=0.2+(X(i)-2)*0.6/4;end
    if (X(i)>=6);pX(i)=0.8;end
end
\end{verbatim}

\noindent\scriptsize\noindent\textbf{sampcost.m}\\[-13pt]
\begin{verbatim}
function qX=sampcost(X);
qX=0.1*ones(size(X));
\end{verbatim}

\section{Matlab Code for Reinforcement Learning Example}
\noindent\normalsize Copy/Paste from the PDF file to the Matlab (or text) editor needs extra work. The symbols `` $^\wedge$ '' and `` $^\prime$ '' used for power and transpose are transferred wrongly. In the editor you need to apply a global replace with the corresponding keyboard symbols.
\vskip0.1cm
\normalsize\noindent Run \texttt{rl\textunderscore main.m} for the example appearing in Fig.\,\ref{fig:3}(a) and (b).
\vskip0.1cm
\noindent\scriptsize\noindent\textbf{rl\textunderscore main.m}\\[-13pt]
\begin{verbatim}
r1=0.8;
r2=0.8;
m1=1;
m2=-1;
s1=1;
s2=1;
gamma=0.8;
rl_num;  % Numerical solution
n=500;
L=100;
times=2000;
I=randi([1 2],2*n,1);
cX=zeros(2*n+1,1);
cX(1)=sqrt(s1)*randn;
for iii=1:2*n
    if I(iii)==1;
        cX(iii+1)=r1*cX(iii)+sqrt(s1)*randn+m1;
    end
    if I(iii)==2;
        cX(iii+1)=r2*cX(iii)+sqrt(s2)*randn+m2;
    end
end
I1=find([I==1]);
X1=cX(I1);
Y1=cX(I1+1);
I2=find([I==2]);
X2=cX(I2);
Y2=cX(I2+1);
R1=reward(Y1);
R2=reward(Y2);
rl_a1   % [A1]
XX=-5+10*[0:500]'/500;
ZZ1=A01*XX'+B01;
XX1=max(ZZ1,0);
WW1=A11'*XX1+B11;
figure(1)
hold on
plot(XX,WW1,'b','linewidth',2)
hold off
ZZ2=A02*XX'+B02;
XX2=max(ZZ2,0);
WW2=A12'*XX2+B12;
figure(1)
hold on
plot(XX,WW2,'color',[0.0 0.6 1],'linewidth',2)
hold off
figure(2)
plot(cost_a11,'b','linewidth',2)
hold on
plot(cost_a12,'color',[0.0 0.6 1],'linewidth',2)
hold off
a=0.2/(1-gamma);
b=1/(1-gamma);
rl_c1  % [C1]
ZZ1=A01*XX'+B01;
XX1=max(ZZ1,0);
WW1=A11'*XX1+B11;
OMWW1=a./(1+exp(WW1))+b./(1+exp(-WW1));
figure(1)
hold on
plot(XX,OMWW1,'r','linewidth',2)
hold off
ZZ2=A02*XX'+B02;
XX2=max(ZZ2,0);
WW2=A12'*XX2+B12;
OMWW2=a./(1+exp(WW2))+b./(1+exp(-WW2));
figure(1)
hold on
plot(XX,OMWW2,'color',[1 0.6 0],'linewidth',2)
hold off
xlabel('$S$','interpreter','latex');
legend('Numerical-1','Numerical-2','[A1]-1','[A1]-2','[C1]-1','[C1]-2','location','north');
figure(2)
hold on
plot(cost_c11,'r')
plot(cost_c12,'color',[1 0.6 0.0]);
xlabel('Number of Interations');
legend('[A1]-1','[A1]-2','[C1]-1','[C1]-2','location','east');
\end{verbatim}

\noindent\scriptsize\noindent\textbf{rl\textunderscore num.m}\\[-13pt]
\begin{verbatim}
times=1000;
X=-20+40*[0:5000]'/5000;
F1=normcdf((X'-r1*X-m1)/sqrt(s1));
[max(F1(:,1)) max(1-F1(:,end))]
F1(:,1)=zeros(length(X),1);F1(:,end)=ones(length(X),1);
F1=[F1(:,2)-F1(:,1) F1(:,3:end)-F1(:,1:end-2) F1(:,end)-F1(:,end-1)]/2;
F2=normcdf((X'-r2*X-m2)/sqrt(s2));
[max(F2(:,1)) max(1-F2(:,end))]
F2(:,1)=zeros(length(X),1);F2(:,end)=ones(length(X),1);
F2=[F2(:,2)-F2(:,1) F2(:,3:end)-F2(:,1:end-2) F2(:,end)-F2(:,end-1)]/2;
U1=ones(size(X));
U2=U1;
R=reward(X);
for iii=1:times
    U1=F1*(R+gamma*max(U1,U2));
    U2=F2*(R+gamma*max(U1,U2));
end
I=[X>=-5]&[X<=5];
figure(1);
plot(X(I),U1(I),'k','linewidth',3);
hold on
plot(X(I),U2(I),'color',[0.6 0.6 0.6],'linewidth',3);
movegui('northwest')
\end{verbatim}

\noindent\scriptsize\noindent\textbf{rl\textunderscore a1.m}\\[-13pt]
\begin{verbatim}
A01=randn(L,1)/sqrt(L);
B01=zeros(L,1);
A11=randn(L,1)/sqrt(L);
B11=0;
A02=randn(L,1)/sqrt(L);
B02=zeros(L,1);
A12=randn(L,1)/sqrt(L);
B12=0;
PDA01=zeros(L,1);
PDB01=zeros(L,1);
PDA11=zeros(L,1);
PDB11=0;
PDA02=zeros(L,1);
PDB02=zeros(L,1);
PDA12=zeros(L,1);
PDB12=0;
mu=0.001;
la=0.99;
c=0.1;
cost_a11=zeros(times,1);
cost_a12=zeros(times,1);
for ttt=1:times
    % Output of the two NN for X1,Y1
    Z11=A01*X1'+B01;
    X11=max(Z11,0);
    W11=A11'*X11+B11;
    OMW11=W11;
    z11=A01*Y1'+B01;
    x11=max(z11,0);
    w11=A11'*x11+B11;
    omw11=w11;
    z12=A02*Y1'+B02;
    x12=max(z12,0);
    w12=A12'*x12+B12;
    omw12=w12;
    YY1=R1+gamma*max(omw11,omw12)';
    cost_a11(ttt)=mean(OMW11.^2/2-YY1'.*OMW11);
    % Computations of derivatives
    U1=-(YY1-OMW11');
    DA11=X11*U1;
    DB11=sum(U1);
    DB01=A11.*(max(sign(Z11),0)*U1);
    DA01=A11.*((max(sign(Z11),0).*X1')*U1);
    if ttt==1
        PDA11=DA11.^2;
        PDB11=DB11.^2;
        PDA01=DA01.^2;
        PDB01=DB01.^2;
    else
        PDA11=la*PDA11+(1-la)*DA11.^2;
        PDB11=la*PDB11+(1-la)*DB11.^2;
        PDA01=la*PDA01+(1-la)*DA01.^2;
        PDB01=la*PDB01+(1-la)*DB01.^2;
    end
    % Update parameters of the first network
    A01=A01-mu*(DA01./sqrt(c+PDA01));
    B01=B01-mu*(DB01./sqrt(c+PDB01));
    A11=A11-mu*(DA11./sqrt(c+PDA11));
    B11=B11-mu*(DB11./sqrt(c+PDB11));
    % Output of the two NN for X2,Y2
    Z22=A02*X2'+B02;
    X22=max(Z22,0);
    W22=A12'*X22+B12;
    OMW22=W22;
    z22=A02*Y2'+B02;
    x22=max(z22,0);
    w22=A12'*x22+B12;
    omw22=w22;
    z21=A01*Y2'+B01;
    x21=max(z21,0);
    w21=A11'*x21+B11;
    omw21=w21;
    YY2=R2+gamma*max(omw21',omw22');
    cost_a12(ttt)=mean(OMW22.^2/2-YY2'.*OMW22);
    % Computations of derivatives
    U2=-(YY2-OMW22');
    DA12=X22*U2;
    DB12=sum(U2);
    DB02=A12.*(max(sign(Z22),0)*U2);
    DA02=A12.*((max(sign(Z22),0).*X2')*U2);
    if ttt==1
        PDA12=DA12.^2;
        PDB12=DB12.^2;
        PDA02=DA02.^2;
        PDB02=DB02.^2;        
    else
        PDA12=la*PDA12+(1-la)*DA12.^2;
        PDB12=la*PDB12+(1-la)*DB12.^2;
        PDA02=la*PDA02+(1-la)*DA02.^2;
        PDB02=la*PDB02+(1-la)*DB02.^2;
    end
    % Update parameters of the second network
    A02=A02-mu*(DA02./sqrt(c+PDA02));
    B02=B02-mu*(DB02./sqrt(c+PDB02));
    A12=A12-mu*(DA12./sqrt(c+PDA12));
    B12=B12-mu*(DB12./sqrt(c+PDB12));
end
\end{verbatim}

\noindent\scriptsize\noindent\textbf{rl\textunderscore c1.m}\\[-14pt]
\begin{verbatim}
A01=randn(L,1)/sqrt(L);
B01=zeros(L,1);
A11=randn(L,1)/sqrt(L);
B11=0;
A02=randn(L,1)/sqrt(L);
B02=zeros(L,1);
A12=randn(L,1)/sqrt(L);
B12=0;
PDA01=zeros(L,1);
PDB01=zeros(L,1);
PDA11=zeros(L,1);
PDB11=0;
PDA02=zeros(L,1);
PDB02=zeros(L,1);
PDA12=zeros(L,1);
PDB12=0;
mu=0.001;
la=0.99;
c=0.1;
cost_c11=zeros(times,1);
cost_c12=zeros(times,1);
for ttt=1:times
    % Output of the two NN for X1,Y1
    Z11=A01*X1'+B01;
    X11=max(Z11,0);
    W11=A11'*X11+B11;
    OMW11=a./(1+exp(W11))+b./(1+exp(-W11));
    z11=A01*Y1'+B01;
    x11=max(z11,0);
    w11=A11'*x11+B11;
    omw11=a./(1+exp(w11))+b./(1+exp(-w11));
    z12=A02*Y1'+B02;
    x12=max(z12,0);
    w12=A12'*x12+B12;
    omw12=a./(1+exp(w12))+b./(1+exp(-w12));
    YY1=R1+gamma*max(omw11,omw12)';
    cost_c11(ttt)=mean((b-a)./(1+exp(W11))+b*log(1+exp(W11))-YY1'.*log(1+exp(W11)));
    % Computation of derivatives
    U1=-(1./(1+exp(-W11'))).*(YY1-OMW11');
    DA11=X11*U1;
    DB11=sum(U1);
    DB01=A11.*(max(sign(Z11),0)*U1);
    DA01=A11.*((max(sign(Z11),0).*X1')*U1);
    if ttt==1
        PDA11=DA11.^2;
        PDB11=DB11.^2;
        PDA01=DA01.^2;
        PDB01=DB01.^2;
    else
        PDA11=la*PDA11+(1-la)*DA11.^2;
        PDB11=la*PDB11+(1-la)*DB11.^2;
        PDA01=la*PDA01+(1-la)*DA01.^2;
        PDB01=la*PDB01+(1-la)*DB01.^2;
    end
    % Update of the first network parameters
    A01=A01-mu*(DA01./sqrt(c+PDA01));
    B01=B01-mu*(DB01./sqrt(c+PDB01));
    A11=A11-mu*(DA11./sqrt(c+PDA11));
    B11=B11-mu*(DB11./sqrt(c+PDB11));
    % Output of the two NN for X2,Y2
    Z22=A02*X2'+B02;
    X22=max(Z22,0);
    W22=A12'*X22+B12;
    z22=A02*Y2'+B02;
    x22=max(z22,0);
    w22=A12'*x22+B12;
    omw22=a./(1+exp(w22))+b./(1+exp(-w22));
    z21=A01*Y2'+B01;
    x21=max(z21,0);
    w21=A11'*x21+B11;
    omw21=a./(1+exp(w21))+b./(1+exp(-w21));
    YY2=R2+gamma*max(omw21',omw22');
    cost_c12(ttt)=mean((b-a)./(1+exp(W22))+b*log(1+exp(W22))-YY2'.*log(1+exp(W22)));
    % Computation of derivatives
    OMW22=a./(1+exp(W22))+b./(1+exp(-W22));
    U2=-(1./(1+exp(-W22'))).*(YY2-OMW22');
    DA12=X22*U2;
    DB12=sum(U2);
    DB02=A12.*(max(sign(Z22),0)*U2);
    DA02=A12.*((max(sign(Z22),0).*X2')*U2);
    if ttt==1
        PDA12=DA12.^2;
        PDB12=DB12.^2;
        PDA02=DA02.^2;
        PDB02=DB02.^2;        
    else
        PDA12=la*PDA12+(1-la)*DA12.^2;
        PDB12=la*PDB12+(1-la)*DB12.^2;
        PDA02=la*PDA02+(1-la)*DA02.^2;
        PDB02=la*PDB02+(1-la)*DB02.^2;
    end
    % Update of the second network parameters
    A02=A02-mu*(DA02./sqrt(c+PDA02));
    B02=B02-mu*(DB02./sqrt(c+PDB02));
    A12=A12-mu*(DA12./sqrt(c+PDA12));
    B12=B12-mu*(DB12./sqrt(c+PDB12));
end
\end{verbatim}

\noindent\scriptsize\noindent\textbf{reward.m}\\[-13pt]
\begin{verbatim}
function R=reward(X);
R=zeros(length(X),1);
for i=1:length(X)
    if X(i)<-7;R(i)=1;end
    if (X(i)>=-7)&(X(i)<-2);R(i)=1-(X(i)+7)*0.8/5;end
    if (X(i)>=-2)&(X(i)<=2);R(i)=0.2;end
    if (X(i)>2)&(X(i)<6);R(i)=0.2+(X(i)-2)*0.6/4;end
    if (X(i)>=6);R(i)=0.8;end
end
\end{verbatim}

\end{document}